\documentclass[sigconf]{acmart}

\newcommand{\hide}[1]{}

\setcopyright{acmlicensed}
\copyrightyear{2024}
\acmYear{2024}
\acmDOI{XXXXXXX.XXXXXXX}
\acmISBN{978-1-4503-XXXX-X/18/06}

\acmConference[SC24 Workshop: AI4S'24]{Workshop on Artificial Intelligence and Machine Learning for Scientific Applications}{November 18, 2024}{Atlanta, GA}

\usepackage{subcaption}

\begin{document}

\title[Active Learning for Data-efficient Surrogate Training]{MelissaDL x Breed: Towards Data-Efficient On-line Supervised Training of Multi-parametric Surrogates with Active Learning}

\author{Sofya Dymchenko}
\email{sofya.dymchenko@inria.fr}
\affiliation{%
	\institution{Univ. Grenoble Alpes, Inria, CNRS, Grenoble INP, LIG}
	\city{Grenoble}
	\country{France}
}

\author{Abhishek Purandare}
\email{abhishek.purandare@inria.fr}
\affiliation{%
	\institution{Univ. Grenoble Alpes, Inria, CNRS, Grenoble INP, LIG}
	\city{Grenoble}
	\country{France}
}

\author{Bruno Raffin}
\email{bruno.raffin@inria.fr}
\affiliation{%
 \institution{Univ. Grenoble Alpes, Inria, CNRS, Grenoble INP, LIG}
 \city{Grenoble}
 \country{France}
}

\begin{abstract}
	
	Artificial intelligence is transforming scientific computing with deep neural network surrogates that approximate solutions to partial differential equations (PDEs). 
	Traditional off-line training methods face issues with storage and I/O efficiency, as the training dataset has to be computed with numerical solvers up-front. 
	Our previous work, the  Melissa framework, addresses these problems by enabling data to be created ``on-the-fly'' and streamed directly into the training process. 
	In this paper we introduce a new active learning method to enhance data-efficiency for on-line  surrogate training.
	The surrogate is direct and multi-parametric, i.e., it is trained to predict a given timestep directly with different initial and boundary conditions parameters. 
	Our approach uses Adaptive Multiple Importance Sampling guided by training loss statistics, in order to focus NN training on the difficult areas of the parameter space.
	Preliminary results for 2D heat PDE demonstrate the potential of this method, called Breed, to improve the generalization capabilities of surrogates while reducing computational overhead.
	
\end{abstract}



\maketitle

\section{Introduction}\label{intro}

The advancement of artificial intelligence (AI) applications in the natural and physical sciences, referred to as \textit{AI4Science (AI4S)}, has noticeably accelerated in recent years~\cite{lavin_simulation_2021}. 
Remarkable examples include molecule docking modeling (AlphaFold~3~\cite{abramson-AccurateStructurePrediction-2024}) and weather forecasting (NeuralGCM~\cite{kochkov-NeuralGeneralCirculation-2024a}, ClimaX~\cite{nguyen-ClimaXFoundationModel-2023}, Aurora~\cite{bodnar2024aurora}).
This progress has been enabled by combining large training data sets, advanced neural architectures, and parallel computational resources. 

Particular attention has been given to \textit{deep surrogates} --- a class of neural networks (NNs) that approximate the solution of partial differential equations (PDEs) used to describe various scientific phenomena.
 PDE solutions are classically obtained using  numerical methods such as Finite Elements Method or Finite Volumes Method. 
 However, these methods are compute and memory intensive. Deep surrogates are expected to deliver quality solutions during inference at a fraction of the memory and computational cost of these solvers.

Deep surrogates can be trained with little to no data, as in Physics Informed NNs (PINNs)~\cite{wang_experts_2023, RAISSI2019686}, where the PDE residuals, initial and boundary conditions (IC, BC) are imposed as soft constraints through the loss. 
Yet, a wide variety of neural architectures, ranging from Graph Neural Networks~\cite{pfaff2020learning} and Neural Operators~\cite{li2020fourier,neurop2024} to Diffusion Models~\cite{kohl-DiffFlow-2023} and Visual Transformers~\cite{herde-PoseidonEfficientFoundation-2024}, are trained with numerical simulations data. 
A surrogate can also be trained in a multi-parametric context with varying ICs and BCs to develop a more generic model. \cite{herde-PoseidonEfficientFoundation-2024} presented a general purpose PDE foundation model that can be finetuned to obtain a specific PDE solution, while \cite{referNO2024} proposed a neural operator architecture that generalizes to unseen geometries.
Since all these NN are trained in a supervised manner --- using data produced by the PDE solver they aim to substitute --- their performance depends on both training set quantity and quality. 

A standard approach is to train  surrogates \textit{off-line}: first, a dataset is generated with traditional PDE solvers and stored on disk; then, the surrogate is trained in an epoch-based manner by reading back the dataset from disk. 
When scaled, this approach suffers from two main limitations: 1) storage capabilities limit the dataset size, thus compromising data quantity, fidelity, and/or diversity; 2) writing and reading the dataset creates an I/O bottleneck, thus impairing efficiency during training. 
In previous works~\cite{meyer-ICML23,meyer-SC23,Schouler-JOSS23}, we tackled both limitations by demonstrating an \textit{on-line} training approach for multi-parametric surrogates with the Melissa framework: the dataset is generated and directly sent to training, bypassing the storage. It allowed us to train a surrogate faster and with higher generalization abilities due to a significantly larger training dataset.
In addition, on-line training enables the steering of the data creation process along the NN training, which opens the way to \textit{active learning (AL)} techniques. AL is a data-centric approach that intends to choose  informative samples to improve data efficiency  for NN training. AL is a  paradigm shift from a model-centric to a data-centric approach, which focuses on the data relevance and quality rather than the model to get better performance~\cite{actdl2020}. \hide{, has been emerging and has already shown advances in many areas of deep learning~\cite{actdl2020}.} 

In this paper, we present our work-in-progress \textit{active learning} method, called Breed, for data-efficient on-line surrogate training with Melissa. To define the input parameters of the next set of solver instances to run, the algorithm relies on the loss values obtained during training and uses Adaptive Importance Sampling method in order to focus on hard-to-learn parameter space regions. We show that for a 2D  Heat PDE, our method chooses initial and boundary conditions with dissimilar temperature values (which are meant to be more challenging to learn by the NN) and, compared to random sampling, the NN overfits noticeably less.

\section{Background}\label{background}

\subsection{Simulation-based deep learning}
Consider partial differential equations (PDE) of the form:
\begin{align}
	\mathcal N[u](x,t) &= f(x,t)  \\
	\mathcal B_i[u](x_i, t)&= g_i(x_i,t) \qquad  \forall x \in \mathcal X, x_i \in \delta\mathcal X_i, t\in\mathcal T\\
	\mathcal I[u](x, 0)&= h(x) 
\end{align}
where $u(x,t)$ is a quantity of interest described by the PDE (e.g., temperature value) defined on a bounded domain $\Omega = \mathcal X \cup \mathcal T \subset R^{d + 1}$. 
$\mathcal N[u]$ is a differential operator acting on $u(x,t)$, 
$\mathcal B_i[u]$ are boundary conditions operators (BCs) for boundaries $\bigcup_i\delta \mathcal X_i =\delta\mathcal X \subset \mathcal X$, and $\mathcal I[u]$ is an initial condition operator (IC). Let  denote  $\lambda$ the vector that encompasses all parameters of PDE (physical constants, coefficients of $f,g,h$).

Most common numerical solvers use mesh-based spatial and temporal discretization and produce trajectories sequentially. To obtain the numerical solution for a considered PDE, we have to 1) provide equation-based functional definitions, domain bounds, and vector $\lambda$; and 2) select spatial discretization size $M^d \cdot \triangle x = \mathcal X$ and temporal discretization size $T \cdot \triangle t = \mathcal T$. The second defines the size of spatial coordinates set $X$ and the number of iterations needed for an autoregressive solver. This not only affects the data fidelity level and approximation quality but also the required memory and computation demands. Let us denote a produced solution field at time step $t=i\cdot\triangle t$ for a given $\lambda_j$ as $x_{ji} = \{ \hat u(x, t) | x\in X\}$, then the solver produces one-by-one a trajectory:
\[
{\lambda_j} : [x_{j,0} \rightarrow x_{j,1} \rightarrow \dots \rightarrow x_{j,T}] \eqcolon x_j
\]

The deep surrogate model $u_{\theta}$ (with weights $\theta$) approximates the PDE solution $u(x,t)$, thereby aiming to substitute a numerical solver. There are different types of surrogate architectures. It can be designed to predict the solution directly, i.e., $u_{\theta}(X, t) = \hat{x}_t$, or autoregressively, i.e., $u_{\theta}(x_t) = \hat{x}_{t+\triangle t}$. As we mentioned in Section~\ref{intro}, the surrogate can be multi-parametric: $u_{\theta}(X, t, \lambda_j) = \hat{x}_j$. \hide{ The surrogates can also be trained  to generalize solutions for different discretizations, as Neural Operators and PINNs, solve an inverse task $u_{\theta}(x^{*}) = \hat{\lambda}$.  for more see Section~\ref{related}.} In the scope of this paper, we consider multi-parametric direct surrogates; details are provided in Section~\ref{exps}.

\subsection{The Melissa DL framework}

Melissa DL~\cite{meyer-ICML23,meyer-SC23, Schouler-JOSS23} is an HPC framework designed for deep learning tasks where the NN is trained with simulation data. It consists of three elements: \textit{clients}, a \textit{server}, and a \textit{launcher} (for details, see Appendix~\ref{app:melissa-details}). 
By default the server uniformly samples input parameters $\lambda$ for each of $S$ clients across the input parameter space $\Lambda$.  Each client then runs the solver on the provided inputs and streams the data (timesteps) to the server. The launcher only submits a subset of all clients based on allocated resources.
This enables the server to dynamically select new input parameters for {\it pending'} clients, which we refer to as \textit{global steering} (hereafter, simply {\it steering}). It is  key to unlocking data-efficient surrogate training with active learning methods.

\section{Active learning steering of data creation for on-line surrogate training}

Active learning  is a possible solution for data-efficient surrogate training, as it can help to reduce the  number of simulations to execute while maintaining the surrogate quality. Generally, AL's goal is to improve NN training by choosing the most informative examples, based on an acquisition function and a query method, to be labeled by an oracle (or a human) and given to the  NN.
In our context, ``labeling by an oracle'' is analogous to ``executing a solver'' and ``choosing examples'' --- to ``sampling solver inputs $\lambda$''. 

However, classical AL methods are not adapted to  our on-line training context.  Extra computational and memory costs imposed by well-known AL techniques are  highly undesirable.  First, the input parameters choice decision has to be fast not to  pause the surrogate training process as the priority is to keep GPUs busy.  Second,  the incurred extra  memory footprint
should be limited to  avoid disk storage  to keep  on-line training efficient. 

To develop AL methods for on-line training, we take inspiration from  methods proposed for data-free PINNs~\cite{wu_comprehensive_2022,daw_mitigating_2023,lau-PINNACLEPINNAdaptive-2023} and extend our previously proposed  method called \textit{Breed}~\cite{myself}.
In our supervised setting, instead of choosing \hide{domain $\Omega$} collocation points, we have to choose input parameters $\lambda$ and run an autoregressive solver. 
In our compute-constrained setting, we aim to use only per-sample loss values as it does not require any extra computation. 
In our memory-constrained setting, we are not able to recompute loss values for all training points but instead have to use ``outdated'' loss values, i.e.  loss values obtained from the  NN at anterior  learning steps.
Additionally, instead of having a pool of points to choose from, we adaptively sample new points based on previous points loss statistics, inspired by Population Monte Carlo algorithms~\cite{cappe-PMC}. We  detail  the proposed method, \textit{Breed}, in the following.

\subsection{Loss-deviation based acquisition metric}

We want to define training sample informativeness through NN loss: the higher the loss, the higher the impact on NN training.
At iteration $i$ of the fixed-state NN $u_{\theta_i}$, there is an underlying probability distribution of NN failure $\mathcal L_{\theta_i}$ over input domain $\Lambda$, which we choose to represent through the self-normalized loss function $L(\cdot,\cdot)$: 
\[
\mathcal L_{\theta_i}(\lambda_{j'}) \approx \frac{ \sum_{t\in \mathcal T} L(u_{\theta_i}(X, t, \lambda_{j'}), x_{j't})}{\sum_{\lambda_j\in\Lambda} \sum_{t\in\mathcal T} L(u_{\theta_i}(X, t, \lambda), x_{jt})},
\]
where $x_{jt}$ is the $t$-th timestep of a trajectory produced by a solver with an input parameter $\lambda_j$, as defined before. 
However we cannot calculate  the ``actual'' per-sample loss values  (corresponding to NN state $\theta_i$)  for neither the whole domain nor approximate them  empirically with all the training points. 
Using per-sample loss values from NN iterations before $i$ is not possible either, as the values are simply not comparable. 
Hence, we propose to approximate it with \textit{per-sample loss deviation statistics} assuming that  the higher the per-sample loss deviation from an average batch loss, the higher the loss.  With this metric points from different batches, hence, different NN states, are comparable. \hide{With this motivation, we give the metric definition in detail.}

Before running the framework, we define the computational budget by choosing a number of total simulations runs $S$, hence, creating a set of input parameters $\Lambda_J = \{\lambda_1, \dots, \lambda_j, \dots \lambda_{S}\}$. 
At any iteration $i'$, there is a set of inputs $\Lambda_J^{(i')} \subseteq \Lambda_J$ of size $S_{\text{done}}$ for which the clients have computed the full trajectories and all points have been seen by the NN. 

Let $l^{(i)}_{jt} = L_{\theta_i}(x_{j,t})$ denote per-sample loss. As a sample can potentially be seen by the NN across several batches $b_i$ before iteration $i'$, we denote the set of these batches indexes $I_{jt}\coloneq \{i | x_{j,t}\in b_{i}\} \subset [0:i']$. Then:
\[
\Lambda_J^{(i')} = \{\lambda_j | \exists l^{(i)}_{jt}, \forall i\in I_{jt}, t=[0:T] \}
\]
We compute and store batch-loss mean $\mu(l^{(i)})$ and standard deviation $\sigma(l^{(i)})$, where $l^{(i)} = \{ l^{(i)}_{jt} | x_{j,t} \in b_i\}$.
Then for any $\lambda_j \in \Lambda_J^{i'}$, we calculate \textit{deviation values} $\delta_{jt} = \{\delta_{jt}^{(i)} | i\in I_{jt}\}$ defined as:
\begin{align}
	\delta ^{(i)}_{jt} = \frac {\max(l_{jt}^{(i)} - \mu(l^{(i)}), 0)}{\sigma(l^{(i)})}
\end{align}
\hide{These per-sample deviation values can be aggregated differently.} We  then  average  across timesteps: 
\begin{align}
	\hat{\mathcal L}_{\theta_{i'}}(\lambda_{j}) = Q_{j} &\coloneq Q(\delta_{jt}) = \frac 1 T  \sum_{t=1:T} \delta_{jt} =\\
	& = \frac 1 T  \sum_{t=1:T} \frac {1}{|I_{jt}|} \sum_{i\in I_{jt}} \delta^{(i)}_{jt}
\end{align}
Not to store all the values, we iteratively update the statistic $\delta_{jt}$ upon the availability of new values.
\hide{ Other aggregation functions $Q$ are possible, like   moving statistics, which is planned for our future work.}

\subsection{Adaptive Multiple Importance Sampling}

Instead of choosing points from a pool or a dataset, we want to sample new points according to progressing $\mathcal L_{\theta}$. We propose to use an Adaptive Multiple Importance Sampling (AMIS) algorithm, inspired by the Population Monte Carlo (PMC) algorithm and previously presented in off-line context~\cite{myself}. An Importance Sampling (IS) goal is to build a proposal probability distribution $q$, which is easy to sample from, to  approximate an unknown target distribution $\pi$, which can be evaluated up to a normalizing constant. 

In PMC, the proposal is built iteratively. At iteration $i$\hide{(not connected to NN iteration from above),}  $q^{(i)}$ is a mixture (population) of $N$ proposals: $q^{(i)} = \sum_{n=1:N} q^{(i)}_n(\cdot|\mu^{(i)}_n, \Sigma)$. The initial locations $\mu^{(0)}_n$ are given or chosen randomly and $\Sigma=\sigma \mathbb I_{d}$ is a hyperparameter. Next, one random value is sampled from each proposal, and an importance weight is calculated:
\begin{align}
	x_n^{(i)} &\sim q^{(i)}_n(\cdot|\mu^{(i)}_n,\sigma) \\
	w_n^{(i)} &= \frac{\pi(x_n^{(i)})}{q_n^{(i)}(x_n^{(i)})}
\end{align}
Then a multinomial distribution with weights $\{w^{(i)}_n\}_{n=1:N}$ is trialed $N$ times, i.e., we resample $\{x^{(i)}_n\}_{n=1:N}$ with replacements and obtain $\{x^{(i)}_{n_i}\}_{n_i\in\{1:N\}, i=1:N}$. The resampled values are used as new location parameters $\mu^{(i+1)}_n$.

\begin{figure}[h!]
	\includegraphics[width=\columnwidth]{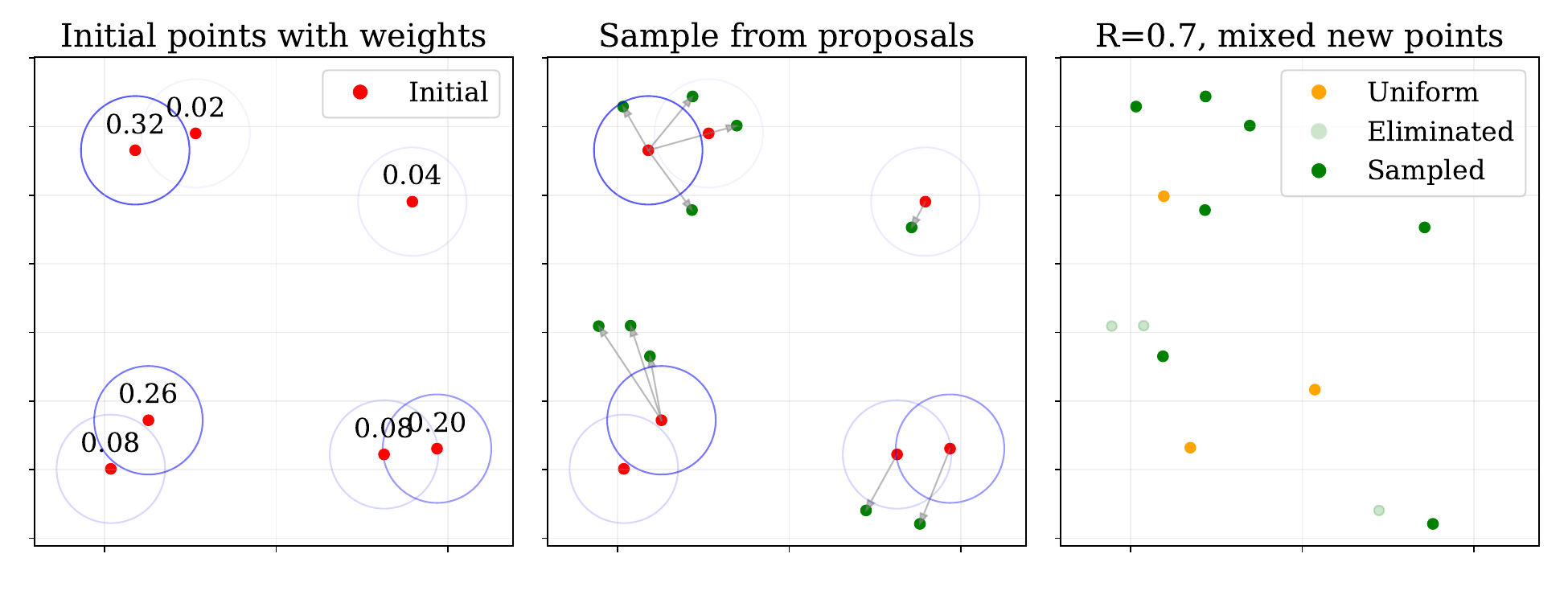}
	\caption{A visual presentation of the sampling algorithm  starting with  $N=7$ initial locations, that are next weighted to build  the Gaussian mixture proposal  and sample from this distribution $K=10$ new samples.
          Last,  30\% of these points are discarded and substituted with uniform points to maintain  exploration capabilities. }
	\label{fig:algo}
\end{figure}

In our context, the target is the approximated distribution $\hat{\mathcal L}_{\theta}$. As obtaining $x_n^{(i)}$ at each PMC iteration for the same $\hat{\mathcal L}_{\theta}$ is not feasible --- NN training is considerably faster than the creation of the data --- we perform only one PMC iteration. We build a proposal once every $P$ NN iterations. Triggering resampling  according to  metrics such  as Effective Sample Size and/or Entropy is left for future work.

At NN iteration $i=0$, the whole set $\Lambda_J$ is sampled uniformly. At resampling iteration $s$, $i=s\cdot P$ (Figure~\ref{fig:algo}), we have a subset of  simulations waiting for execution, i.e. $\tilde{\Lambda}_J^{(i)} \coloneq \Lambda_J \setminus \Lambda_J^{(i)}$ of size $K=S-S_{\text{done}}$. We want to substitute the corresponding input parameters by sampling $K$ new points. To build a proposal $q^{(s)}$, we use a population of window size $N\leq S_{\text{done}}$, i.e., last $\lambda_j \in \Lambda_J^{(i)}$ in order of $Q_j$ value updates,  for which we keep the notation  $\Lambda_J^{(i)}$. Here, the points $\lambda_j$ are the initial locations of the proposal, and $Q_j$ are target distribution evaluations. Then importance weights\footnote{Normally, we should divide by a proposal likelihood, but in experiments, we have not noticed if division affects the quality, so it is omitted in this paper.} are:
\begin{align}
	w_{j'} = \frac{Q_{j'}}{\frac 1 N \sum_{j} Q_j} \quad \forall \lambda_j', \lambda_j \in \Lambda_J^{(i)}
\end{align}
and we resample $k\!=\!1\!:\!K$ locations to build the proposal:
\begin{align}
	\lambda_{j_k} &\sim \text{Mult}(\Lambda_J^{(i)}, w_j, K)\\
	q^{(s)}(\cdot) = \sum_k q^{(s)}_{k}(\cdot) &= \sum_k \text{Gauss} (\cdot|\lambda_{j_k}, \sigma).
\end{align}
Finally, we resample new input parameters:
\begin{align}
	\tilde{\Lambda}_J^{(s)} \leftarrow \{\lambda_k \sim q^{(s)}_k(\cdot)\}_k \label{obtain}
\end{align}

In our implementation, the complexity of one iteration is $O(K)$, though it can be parallelized. If the point sampled appeared out of bounds, we decrease $\sigma$ by $3e^{-1}$ for its proposal member and sample again. We do it at most five times, and otherwise, the location is left the same. The decreased $\sigma$ is passed to this proposal member. We use \verb|MultivariateNormal| class from \verb|Pytorch|.

The IS algorithms are known to suffer from mode collapse and underexploration. To tackle this issue and balance a training set, we create \textit{a mixture distribution}: $r^{(s)} \cdot q^{(s)}(\cdot) + (1 - r^{(s)}) \cdot \mathcal U(\Lambda)$. In our implementation (Eq.~\eqref{obtain}), inputs are substituted with uniform points with probability $r^{(s)}$. \textit{The concentrate-explore value $r^{(s)}\in[0,1]$} changes as: $r^{(s)} = \max\left(s\cdot\frac{r_e-r_s}{r_c}, r_e\right)$. The triplet $(r_s, r_e, r_c)$ is a hyperparameter. 

\subsection{HPC implementation details}
\begin{figure}[h!]
	\centering
	\includegraphics[width=0.35\textwidth]{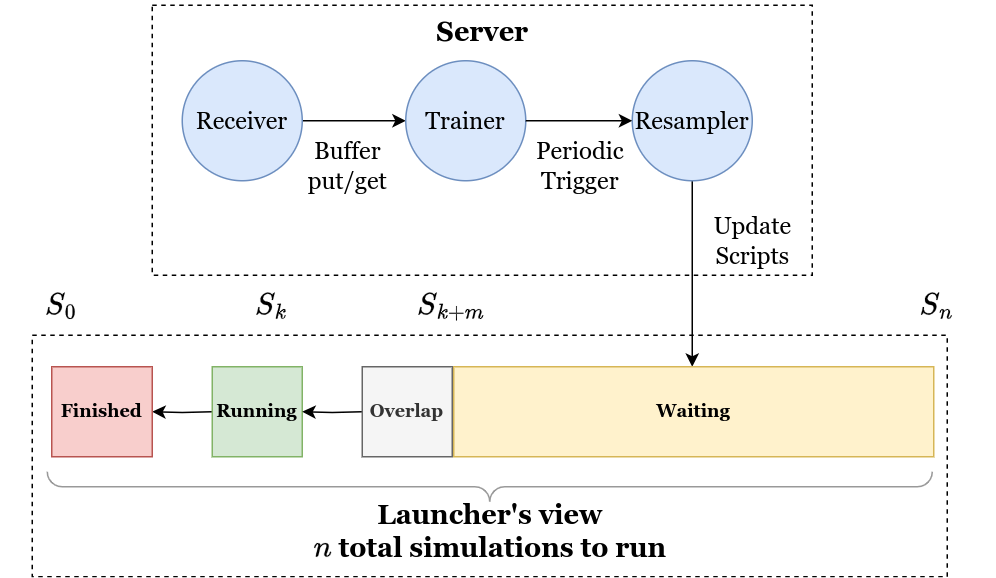}
	\caption{The server's communication with the launcher for the input parameters update mechanism. The number of simulations to run is defined by the budget $n$. \hide{Whe triggering resampling, the number of already executed jobs as well as the waiting one is queried to the server. }}
	\label{fig:global-steering-launcher-server}
\end{figure}

We expand the Melissa DL server with the steering mechanism (Figure~\ref{fig:global-steering-launcher-server}) to apply Breed. The resampling is triggered by the server periodically based on the NN training iteration. Firstly, the server acquires a consistent view of the launcher's job submissions. Secondly, it identifies the simulations that have not yet been submitted for resampling the inputs and, finally, starts the resampling algorithm.

The Melissa launcher has a limit $m$ on the maximum number of jobs allowed to run simultaneously, determined by the available resources. 
Assuming the trigger is invoked while currently running or submitted simulation $S_k$, where $k$ is the highest simulation ID observed from the launcher's perspective. The exact start time of the next simulations from $S_{k+1}$ to $S_{k+m-1}$ cannot be determined due to the inherent uncertainty of the batch scheduler. Therefore, the server always chooses $S_{k+m}$ as the starting point and thus avoids inconsistencies that lead to resampling the parameters that may have been already submitted.

The primary limitation of this approach lies in selecting the appropriate trigger period. 
For instance, if the period is too frequent, a resampled generation might never execute with the same parameters as it is likely to be overwritten multiple times. 
Consequently, this value is left to the user's discretion, taking into account the execution speeds of both the solvers and the training process. \hide{, as well as other relevant hyperparameters.}

\section{Experimental study}\label{exps}

We experiment with a 2D Heat PDE solver called HeatPDE (Appendix~\ref{app:general-setup-heatpde}), focusing on the analysis of the method's hyperparameters space and its performance with different NN sizes. We compare our method to an on-line training of a surrogate where input parameters are sampled uniformly, which we refer to as \textit{Random}. We chose the heat PDE case due to its relatively low computational demands and ease of interpretability.

The surrogate is trained to directly predict the discretized temperature field: $u_{\theta}(\lambda, t) = \hat{u}_{\lambda}(x,t)$, where $\lambda=[T_0, T_1, T_2, T_3, T_4] \in [100,500]^5\subset\Lambda$ are the initial and four boundary temperatures and $t\in [0,1,\dots,100]$. We choose a multilayer perceptron with an input layer of 6 neurons, $L$ hidden layers of $H$ neurons with ReLU activation, and an output of $M^2=64^2$ neurons. It is trained using Adam optimizer with a learning rate of $1e^{-3}$ and batch size $B=128$. 
The simulations run budget is $S=800$, and the pre-created fixed validation set has 200 full-trajectory simulations with parameters generated from a  quasi-uniform Halton sequence. \hide{, which ensures a good coverage of $\Lambda$.}

\subsection{Study description}

\begin{figure*}[ht!]
	\begin{tabular}[b]{cc}
		\begin{subfigure}[b]{0.35\textwidth}
			\includegraphics[width=\textwidth]{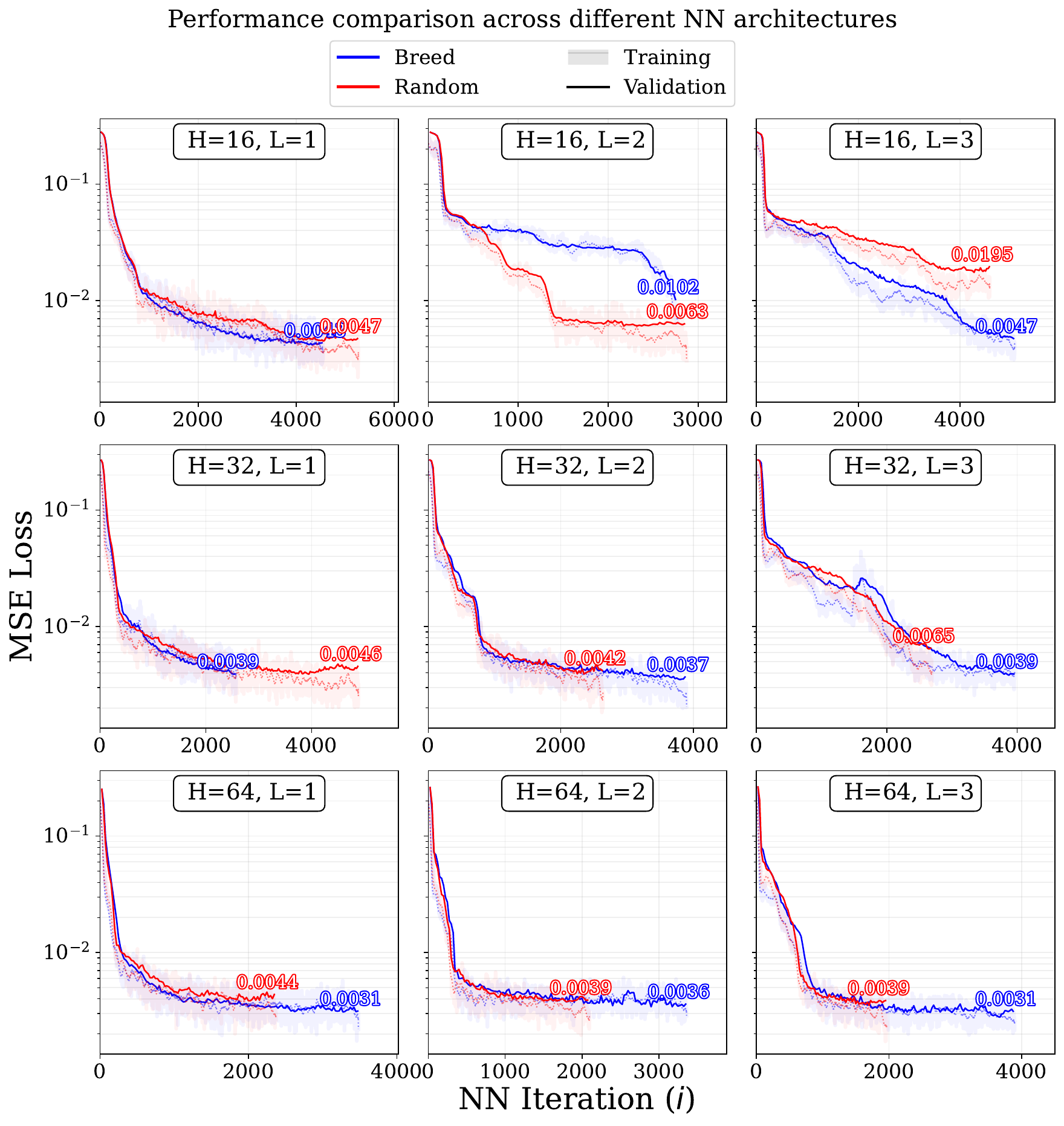}
			\caption{Comparative study of models: each plot represents a run with varying fully connected NN configurations, i.e., hidden layer size $H$ (different rows) and number of layers $L$ (different columns). }
			\label{fig:modelchange}
		\end{subfigure}
		&
		\begin{subfigure}[b]{0.65\textwidth}
			\includegraphics[width=\textwidth]{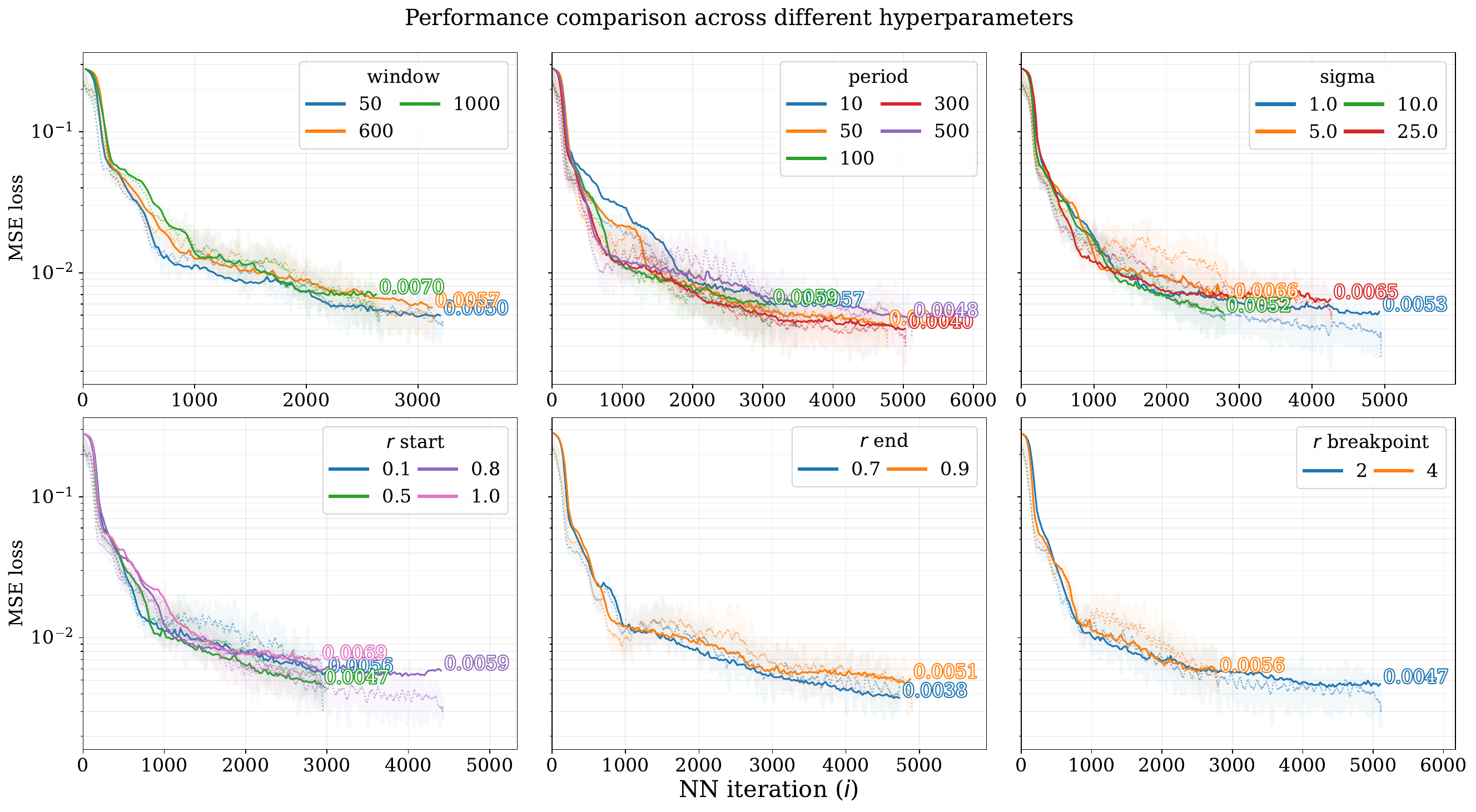}
			\vspace{1em}
			\caption{Comparative study of Breed: each plot represents a run with one varying hyperparameter value, while every other is fixed, including model configuration.}
			\label{fig:parameterchange}
		\end{subfigure}
	\end{tabular}
	\label{fig:change}
	\caption{Experimental study over hyperparameters. The changing parameter is indicated in each legend box. The training curve is averaged with a moving window of 40 iterations (dotted line) for visibility. The Y-axis is a logarithmic scale, and it is shared across all plots. Values presented near the curves are the last validation loss values. \hide{See Appendix~\ref{app:hptable} for a detailed table of fixed values.}}
\end{figure*}

We conduct two systematic studies: across different model configurations (Figure~\ref{fig:modelchange}) and across different Breed hyperparameters (Figure~\ref{fig:parameterchange}). All other configuration values are fixed for fair comparison (see appendix~Table~\ref{tab:fixed}). We vary:
\begin{enumerate}
	\item The hidden size $H$ and number of layers $L$ of the fully connected NN;
	\item The 3  parameters associated with sampling: window size $N$ and period $P$ (implementation), width $\sigma$ (PMC);
	\item The 3 parameters associated with $r$ value: $(r_s, r_e, r_c)$.
\end{enumerate}

The model configuration affects its expressivity and capability to capture more complex data. It directly connects to overfitting and underfitting phenomena, which in an on-line setting are specifically important. We study values $H=[16,32,64]$ and $L=[1,2,3]$, run two experiments, with Random and Breed steering.

The window size defines the size of the proposal population, which might affect distribution approximation: smaller values can make it ``unstable'' while bigger values can make it ``outdated''. We study values $[50, 600, 1000]$.

The period affects the computational load --- how often we perform resampling \hide{(which potentially can create an overhead job-submissions-wise)}.  It  can also  affect the distribution approximation quality as Breed is trying to follow  a dynamically changing target $\mathcal L_{\theta_i}$. \hide{, the quality depends on NN convergence speed. } Currently, the period is static, but  we expect  to extend it to an adaptive trigger that uses the usual MCMC modeling metrics, e.g., effective size and expected improvement. We study values $[10, 50, 100, 300, 500]$.

The tuning of width $\sigma$ is a known issue in PMC algorithms. Smaller values might make the sampling too ``myopic'' while bigger values might make it not concentrative enough. Finding a golden middle is challenging. We study values $[1.0, 5.0, 10.0, 25.0]$.

The biggest tuning burden is created by $r$-value, as it is specific to the problem, the model, and parameter $P$. However, this mixing ratio was the simplest technique for the exploitation-exploration dilemma, which also appears in MCMC modeling and reinforcement learning. We chose a ``linear-constant'' change scheme  based on heuristics from our previous work, where we noticed that a ``warming up'' period was needed to prevent convergence destabilization. In future work, our main goal is to adopt adaptive solutions. We study values $r_s=[0.1,0.5,0.8,1.0], r_e=[0.7, 0.9], r_c=[2, 4]$. 

\subsection{Results discussion}

While overall Breed performance is not clearly better than Random, which can be explained by HeatPDE case simplicity, we see a specific pattern. Given higher expressivity to the model, Random experiments tend to show overfitting, which is especially noticeable for $H=16$, $L=3$, while \textit{Breed training and validation curves stay close \footnote{In Melissa, a training thread may operate more frequently than a receiving thread. It can result in more training iterations independent of the run configuration, which we observe in figures.}} (Figure~\ref{fig:modelchange}). 

We see overfitting for some hyperparameter values (Figure~\ref{fig:parameterchange}), i.e., high $r_s$ and low $\sigma$. As for the convergence, we observe that higher window sizes and lower periods tend to make training divergent at the beginning, which affects further iterations.  Within the $r_s$ study, we see that value $0.5$ also shows divergence at earlier iterations, but at later ones, converges faster, and its training loss is higher than validation loss, which is a good sign of generalization abilities. Consequently, $r_e$ and $r_c$ affect the training as well.

\begin{figure}[h!]
	\begin{subfigure}[b]{0.47\columnwidth}
		\includegraphics[width=1.1\textwidth]{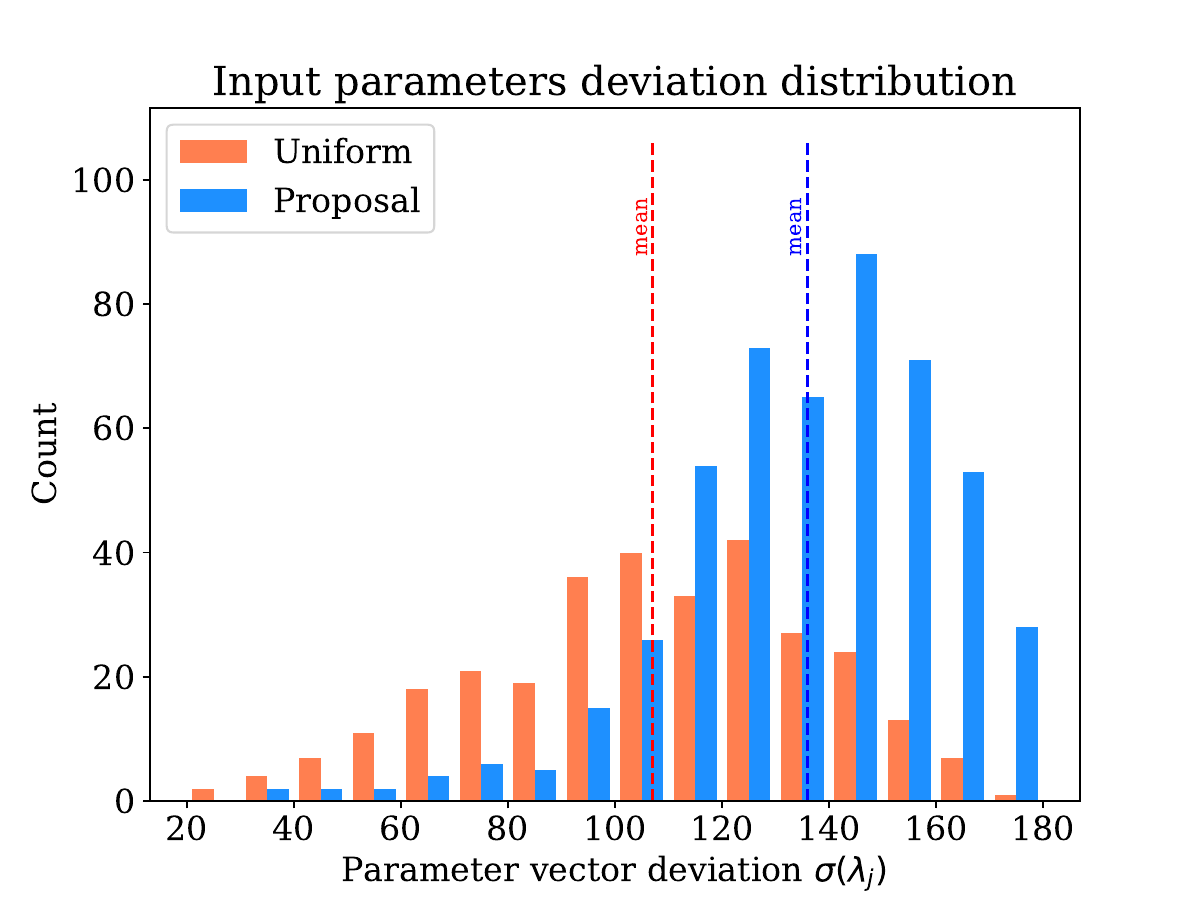}
		\caption{Breed run: uniform (orange) against proposal (light blue)}
		\label{fig:unifprop}
	\end{subfigure}
	\hfill
	\begin{subfigure}[b]{0.47\columnwidth}
		\includegraphics[width=1.1\textwidth]{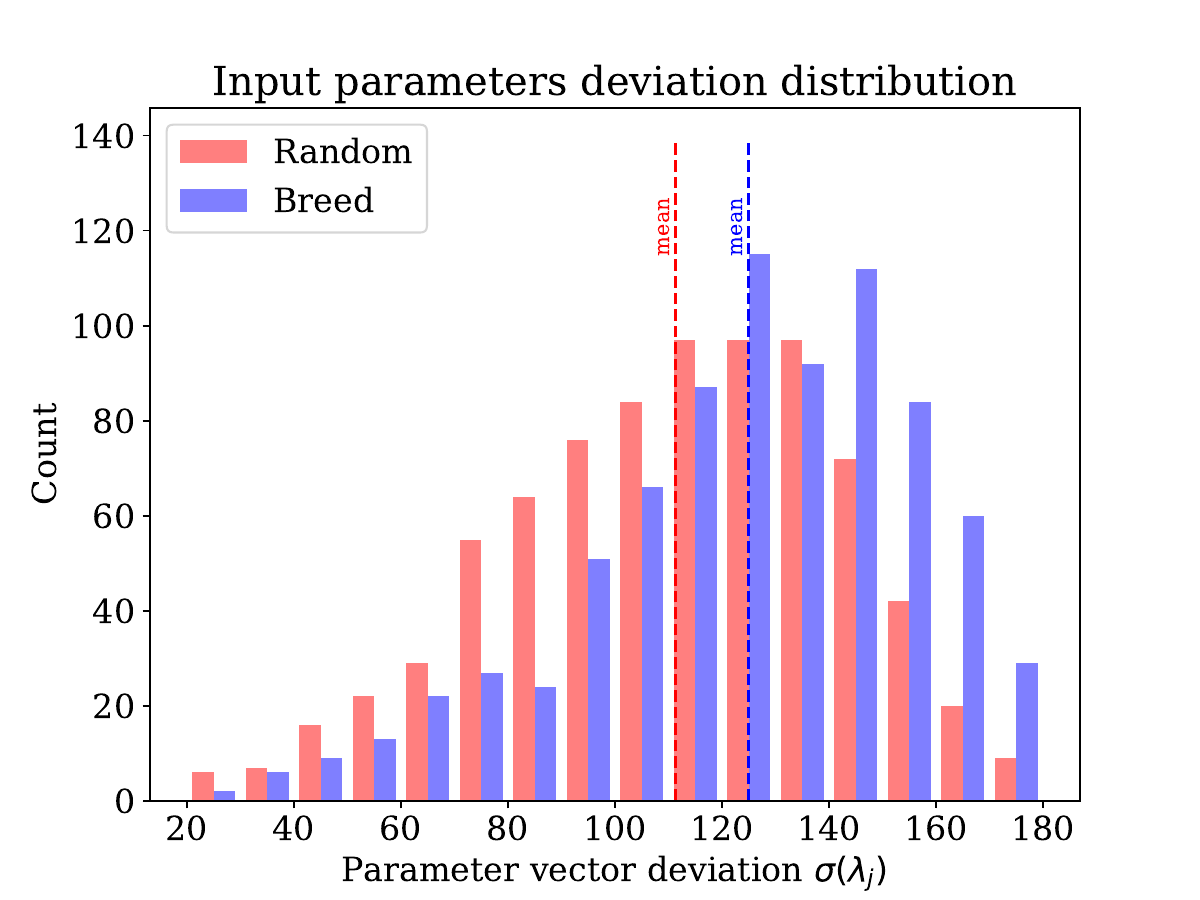}
		\caption{Random run (red) against Breed run (blue)}
		\label{fig:randbreed}
	\end{subfigure}
	\caption{Input parameter deviation histogram  obtained from one run of  800 input parameters. On the left (a), comparison  per source of point (whether uniform or proposal) for one Breed run; on the right (b), comparison of two runs (Random and Breed). The mean is plotted to better see the distribution shift.}
        \label{fig:distributionshift}
\end{figure}

Apart from performance analysis, we conducted explorative analysis across training statistics. Our central insight is that  the conditional distribution of input parameters created overall for the run is clearly shifted when we use Breed (Figure~\ref{fig:distributionshift}). We calculated a per  input vector deviation, which represents how large is the difference between the temperatures $T_{0:5}$, and built a histogram. In Figure~\ref{fig:randbreed}, we compare the fixed configuration run with the Random and Breed methods. The mean of the latter distribution is shifted toward higher deviation values. It means that \textit{Breed focuses sampling in $\Lambda$ regions where temperatures are more diverse}. It makes sense for the HeatPDE case, as diverse temperatures bring more dynamicity to the trajectory, which should be harder to learn. To see this phenomenon clearly, we compared the  histograms for the uniform and proposal samples for  one Breed run in Figure~\ref{fig:unifprop}. 

Additionally, we calculated the correlation coefficients between the NN iteration, per-sample and batch losses, and our proposed deviation loss metric $Q_j$. We noticed that our metric has no correlation to the  NN iteration (-0.02) but has a positive correlation (0.27) with per-sample loss, while batch loss and sample loss correlate with the NN iteration (0.4, 0.31). It means, \textit{we constructed a metric that is comparable between NN iterations and partially representative of per-sample loss}. See a visual representation of the correlation matrix in appendix Figure~\ref{fig:corr}.

\section{Related work}\label{related}

The question of active learning originated in the context of training NNs with a finite dataset~\cite{Takezoe-DeepActiveLearning-2023}. The goal is to ``select a small subset of unlabeled samples from a large pool of data for labeling and training, while achieving comparable generalization performance to learning on the entire dataset''~\cite{bu-ProvablyNeuralActive-2024}. Active learning in that context relies on two main criteria. The first one is based on \textit{uncertainty}, choosing samples that the neural models are most uncertain about. The second is based on \textit{diversity}, selecting samples bringing diversity in the feature space compared to the already labeled ones~\cite{bu-ProvablyNeuralActive-2024}.
It is tackled in various ways: measuring samples uncertainty by approximating training dynamics~\cite{kye_tidal_2022, wang_deep_2022}, calculating samples influence~\cite{k_revisiting_2021}, selecting representative subsets with use of gradients~\cite{killamsetty_automata_2022, fayyaz_bert_2022, katharopoulos_not_2019}, with also extensions to data streams~\cite{cacciarelli-ActiveLearningData-2024}. 

In the AI4Science domain, active learning has recently seen a surge of interest to improve PINNs training.
PINNs are trained by choosing collocation points. Uniform sampling is the standard approach, but alternative adaptive sampling strategies have been proposed in an attempt to improve noticeably hard-to-train networks.
Approach ranges from re-weighting sample importance~\cite{yang_dmis_2022, nabian_efficient_2021},
creating a training subset based on a probability distribution calculated from  the normalized losses~\cite{wu_comprehensive_2022}, or retaining points whose loss is higher than average and resampling the remaining ones uniformly for each iteration~\cite{daw_mitigating_2023}. \cite{lau-PINNACLEPINNAdaptive-2023} proposes to use values calculated with the Neural Tangent Kernel (NTK) instead of the loss, giving a more precise insight into the influence of each sample.  The compute overhead is shown to be compensated by the gain on the convergence speed. 

The paper~\cite{musekamp-ActiveLearningNeural-2024} proposes to apply classical active learning algorithms to multi-parametric surrogate training. Their approach is fully off-line, using active learning with lightweight NNs to gather metrics to build a static data set, which is next used for training the full-featured surrogate. Simulation-based inference (SBI)~\cite{cranmer-FrontierSimulationBasedInference-2020} trains a NN from simulations to solve an inverse problem. The NN often relies on normalizing flows to learn the posterior distribution, which can be used in turn to select the input parameters of the next set of simulations to run. We can also mention a former work~\cite{Baydin2019} that addressed a similar inference issue from simulation runs, using an LSTM neural architecture with the particularity of experiments at large scale on more than thousands of CPU nodes. Another work proposed an HPC framework that relies on machine learning to adaptively select the members of an ensemble to run to accelerate the parameter space exploration according to some given objective~\cite{ward_colmena_2021}.

Our proposed sampling method is inspired by a PMC algorithm. To our knowledge, PMC has not been used for active learning. 
More advanced PMC versions exist that instead of sampling within the vicinity of high probability points with a fixed standard deviation, exploit geometric information of the target to adapt the location and scale parameters of those proposals~\cite{elvira-OptimizedPopulationMonte-2022, elvira-GradientbasedAdaptiveImportance-2023} or use normalizing flows instead of a Gaussian proposal~\cite{pal_population_2023}. Deploying such an algorithm in our context is left for future work. 

\section{Conclusion}

In this paper, we introduced an active learning method for data-efficient on-line training of deep surrogates  using the Melissa framework. Our approach steers data creation by leveraging loss statistics and Importance Sampling: it guides the solvers to compute trajectories with input parameters in hard-to-learn regions. Preliminary results \hide{from a broad experimental study for} with  the heat equation showed Breed's potential to improve NN generalization ability without computational overhead, as well as an interpretable choice of points. \hide{Overall, this paper represents a step forward in addressing the limitations of traditional surrogate training methods and opens new avenues for developing more efficient and capable AI-driven scientific models.} Future work will focus on conducting experiments for larger-scale and more complex dynamic PDEs, refining the method with advanced sampling techniques, and reducing the set of hyperparameters by developing self-adaptive techniques.

\section*{Acknowledgements}
	
	This work was supported by project Exa-DoST, NumPEx PEPR program, France 2030 state grant reference ANR-22-EXNU-0004. This work was granted access to the HPC resources of IDRIS under the allocations AD010610366R3 attributed by GENCI (Grand Equipement National de Calcul Intensif), and benefited from access to the Grid'5000 testbed, supported by a scientific interest group hosted by Inria and including CNRS, RENATER and several Universities as well as other organizations. We are  thankful to Quentin Guilloteau and Fernando Ayats for their help with setting the environment for reproducible experiments.

\bibliographystyle{ACM-Reference-Format}
\bibliography{bibfile}


\begin{thebibliography}{44}


\ifx \showCODEN    \undefined \def \showCODEN     #1{\unskip}     \fi
\ifx \showDOI      \undefined \def \showDOI       #1{#1}\fi
\ifx \showISBNx    \undefined \def \showISBNx     #1{\unskip}     \fi
\ifx \showISBNxiii \undefined \def \showISBNxiii  #1{\unskip}     \fi
\ifx \showISSN     \undefined \def \showISSN      #1{\unskip}     \fi
\ifx \showLCCN     \undefined \def \showLCCN      #1{\unskip}     \fi
\ifx \shownote     \undefined \def \shownote      #1{#1}          \fi
\ifx \showarticletitle \undefined \def \showarticletitle #1{#1}   \fi
\ifx \showURL      \undefined \def \showURL       {\relax}        \fi
\providecommand\bibfield[2]{#2}
\providecommand\bibinfo[2]{#2}
\providecommand\natexlab[1]{#1}
\providecommand\showeprint[2][]{arXiv:#2}

\bibitem[\protect\citeauthoryear{Abramson, Adler, Dunger, Evans, Green,
  Pritzel, Ronneberger, Willmore, Ballard, Bambrick, Bodenstein, Evans, Hung,
  O'Neill, Reiman, Tunyasuvunakool, Wu, {\v Z}emgulyt{\.e}, Arvaniti, Beattie,
  Bertolli, Bridgland, Cherepanov, Congreve, {Cowen-Rivers}, Cowie, Figurnov,
  Fuchs, Gladman, Jain, Khan, Low, Perlin, Potapenko, Savy, Singh, Stecula,
  Thillaisundaram, Tong, Yakneen, Zhong, Zielinski, {\v Z}{\'i}dek, Bapst,
  Kohli, Jaderberg, Hassabis, and Jumper}{Abramson et~al\mbox{.}}{2024}]%
        {abramson-AccurateStructurePrediction-2024}
\bibfield{author}{\bibinfo{person}{Josh Abramson}, \bibinfo{person}{Jonas
  Adler}, \bibinfo{person}{Jack Dunger}, \bibinfo{person}{Richard Evans},
  \bibinfo{person}{Tim Green}, \bibinfo{person}{Alexander Pritzel},
  \bibinfo{person}{Olaf Ronneberger}, \bibinfo{person}{Lindsay Willmore},
  \bibinfo{person}{Andrew~J. Ballard}, \bibinfo{person}{Joshua Bambrick},
  \bibinfo{person}{Sebastian~W. Bodenstein}, \bibinfo{person}{David~A. Evans},
  \bibinfo{person}{Chia-Chun Hung}, \bibinfo{person}{Michael O'Neill},
  \bibinfo{person}{David Reiman}, \bibinfo{person}{Kathryn Tunyasuvunakool},
  \bibinfo{person}{Zachary Wu}, \bibinfo{person}{Akvil{\.e} {\v
  Z}emgulyt{\.e}}, \bibinfo{person}{Eirini Arvaniti}, \bibinfo{person}{Charles
  Beattie}, \bibinfo{person}{Ottavia Bertolli}, \bibinfo{person}{Alex
  Bridgland}, \bibinfo{person}{Alexey Cherepanov}, \bibinfo{person}{Miles
  Congreve}, \bibinfo{person}{Alexander~I. {Cowen-Rivers}},
  \bibinfo{person}{Andrew Cowie}, \bibinfo{person}{Michael Figurnov},
  \bibinfo{person}{Fabian~B. Fuchs}, \bibinfo{person}{Hannah Gladman},
  \bibinfo{person}{Rishub Jain}, \bibinfo{person}{Yousuf~A. Khan},
  \bibinfo{person}{Caroline M.~R. Low}, \bibinfo{person}{Kuba Perlin},
  \bibinfo{person}{Anna Potapenko}, \bibinfo{person}{Pascal Savy},
  \bibinfo{person}{Sukhdeep Singh}, \bibinfo{person}{Adrian Stecula},
  \bibinfo{person}{Ashok Thillaisundaram}, \bibinfo{person}{Catherine Tong},
  \bibinfo{person}{Sergei Yakneen}, \bibinfo{person}{Ellen~D. Zhong},
  \bibinfo{person}{Michal Zielinski}, \bibinfo{person}{Augustin {\v
  Z}{\'i}dek}, \bibinfo{person}{Victor Bapst}, \bibinfo{person}{Pushmeet
  Kohli}, \bibinfo{person}{Max Jaderberg}, \bibinfo{person}{Demis Hassabis},
  {and} \bibinfo{person}{John~M. Jumper}.} \bibinfo{year}{2024}\natexlab{}.
\newblock \showarticletitle{Accurate Structure Prediction of Biomolecular
  Interactions with {{AlphaFold}} 3}.
\newblock \bibinfo{journal}{\emph{Nature}} \bibinfo{volume}{630},
  \bibinfo{number}{8016} (\bibinfo{date}{June} \bibinfo{year}{2024}),
  \bibinfo{pages}{493--500}.
\newblock
\showISSN{1476-4687}
\urldef\tempurl%
\url{https://doi.org/10.1038/s41586-024-07487-w}
\showDOI{\tempurl}


\bibitem[\protect\citeauthoryear{{Azizzadenesheli}, {Kovachki}, {Li},
  {Liu-Schiaffini}, {Kossaifi}, and {Anandkumar}}{{Azizzadenesheli}
  et~al\mbox{.}}{2024}]%
        {neurop2024}
\bibfield{author}{\bibinfo{person}{Kamyar {Azizzadenesheli}},
  \bibinfo{person}{Nikola {Kovachki}}, \bibinfo{person}{Zongyi {Li}},
  \bibinfo{person}{Miguel {Liu-Schiaffini}}, \bibinfo{person}{Jean {Kossaifi}},
  {and} \bibinfo{person}{Anima {Anandkumar}}.} \bibinfo{year}{2024}\natexlab{}.
\newblock \showarticletitle{{Neural operators for accelerating scientific
  simulations and design}}.
\newblock \bibinfo{journal}{\emph{Nature Reviews Physics}} \bibinfo{volume}{6},
  \bibinfo{number}{5} (\bibinfo{date}{May} \bibinfo{year}{2024}),
  \bibinfo{pages}{320--328}.
\newblock
\urldef\tempurl%
\url{https://doi.org/10.1038/s42254-024-00712-5}
\showDOI{\tempurl}
\showeprint[arxiv]{2309.15325}~[cs.LG]


\bibitem[\protect\citeauthoryear{Balouek, Amarie, Charrier, Desprez, Jeannot,
  Jeanvoine, L{\`e}bre, Margery, Niclausse, Nussbaum, et~al\mbox{.}}{Balouek
  et~al\mbox{.}}{2013}]%
        {grid}
\bibfield{author}{\bibinfo{person}{Daniel Balouek},
  \bibinfo{person}{Alexandra~Carpen Amarie}, \bibinfo{person}{Ghislain
  Charrier}, \bibinfo{person}{Fr{\'e}d{\'e}ric Desprez},
  \bibinfo{person}{Emmanuel Jeannot}, \bibinfo{person}{Emmanuel Jeanvoine},
  \bibinfo{person}{Adrien L{\`e}bre}, \bibinfo{person}{David Margery},
  \bibinfo{person}{Nicolas Niclausse}, \bibinfo{person}{Lucas Nussbaum},
  {et~al\mbox{.}}} \bibinfo{year}{2013}\natexlab{}.
\newblock \showarticletitle{Adding virtualization capabilities to the
  Grid’5000 testbed}. In \bibinfo{booktitle}{\emph{Cloud Computing and
  Services Science: Second International Conference, CLOSER 2012, Porto,
  Portugal, April 18-21, 2012. Revised Selected Papers 2}}. Springer,
  \bibinfo{pages}{3--20}.
\newblock


\bibitem[\protect\citeauthoryear{Baydin, Shao, Bhimji, Heinrich, Meadows, Liu,
  Munk, Naderiparizi, Gram-Hansen, Louppe, Ma, Zhao, Torr, Lee, Cranmer,
  {Prabhat}, and Wood}{Baydin et~al\mbox{.}}{2019}]%
        {Baydin2019}
\bibfield{author}{\bibinfo{person}{Atılım~Güneş Baydin},
  \bibinfo{person}{Lei Shao}, \bibinfo{person}{Wahid Bhimji},
  \bibinfo{person}{Lukas Heinrich}, \bibinfo{person}{Lawrence Meadows},
  \bibinfo{person}{Jialin Liu}, \bibinfo{person}{Andreas Munk},
  \bibinfo{person}{Saeid Naderiparizi}, \bibinfo{person}{Bradley Gram-Hansen},
  \bibinfo{person}{Gilles Louppe}, \bibinfo{person}{Mingfei Ma},
  \bibinfo{person}{Xiaohui Zhao}, \bibinfo{person}{Philip Torr},
  \bibinfo{person}{Victor Lee}, \bibinfo{person}{Kyle Cranmer},
  \bibinfo{person}{{Prabhat}}, {and} \bibinfo{person}{Frank Wood}.}
  \bibinfo{year}{2019}\natexlab{}.
\newblock \showarticletitle{Etalumis: Bringing Probabilistic Programming to
  Scientific Simulators at Scale}.
\newblock  (\bibinfo{year}{2019}).
\newblock
\urldef\tempurl%
\url{https://doi.org/10.1145/3295500.3356180}
\showDOI{\tempurl}
\showeprint[arxiv]{1907.03382}
\newblock
\shownote{Publisher: {IEEE} Computer Society.}


\bibitem[\protect\citeauthoryear{Bodnar, Bruinsma, Lucic, Stanley,
  Brandstetter, Garvan, Riechert, Weyn, Dong, Vaughan, Gupta, Thambiratnam,
  Archibald, Heider, Welling, Turner, and Perdikaris}{Bodnar
  et~al\mbox{.}}{2024}]%
        {bodnar2024aurora}
\bibfield{author}{\bibinfo{person}{Cristian Bodnar}, \bibinfo{person}{Wessel
  Bruinsma}, \bibinfo{person}{Ana Lucic}, \bibinfo{person}{Megan Stanley},
  \bibinfo{person}{Johannes Brandstetter}, \bibinfo{person}{Patrick Garvan},
  \bibinfo{person}{Maik Riechert}, \bibinfo{person}{Jonathan Weyn},
  \bibinfo{person}{Haiyu Dong}, \bibinfo{person}{Anna Vaughan},
  \bibinfo{person}{Jayesh Gupta}, \bibinfo{person}{Kit Thambiratnam},
  \bibinfo{person}{Alex Archibald}, \bibinfo{person}{Elizabeth Heider},
  \bibinfo{person}{Max Welling}, \bibinfo{person}{Richard Turner}, {and}
  \bibinfo{person}{Paris Perdikaris}.} \bibinfo{year}{2024}\natexlab{}.
\newblock \bibinfo{booktitle}{\emph{Aurora: A Foundation Model of the
  Atmosphere}}.
\newblock \bibinfo{type}{{T}echnical {R}eport} MSR-TR-2024-16.
  \bibinfo{institution}{Microsoft Research AI for Science}.
\newblock
\urldef\tempurl%
\url{https://www.microsoft.com/en-us/research/publication/aurora-a-foundation-model-of-the-atmosphere/}
\showURL{%
\tempurl}


\bibitem[\protect\citeauthoryear{Bu, Huang, Suzuki, Cheng, Zhang, Xu, and
  Wong}{Bu et~al\mbox{.}}{2024}]%
        {bu-ProvablyNeuralActive-2024}
\bibfield{author}{\bibinfo{person}{Dake Bu}, \bibinfo{person}{Wei Huang},
  \bibinfo{person}{Taiji Suzuki}, \bibinfo{person}{Ji Cheng},
  \bibinfo{person}{Qingfu Zhang}, \bibinfo{person}{Zhiqiang Xu}, {and}
  \bibinfo{person}{Hau-San Wong}.} \bibinfo{year}{2024}\natexlab{}.
\newblock \showarticletitle{Provably {{Neural Active Learning Succeeds}} via
  {{Prioritizing Perplexing Samples}}}. In
  \bibinfo{booktitle}{\emph{Forty-First {{International Conference}} on
  {{Machine Learning}}}}.
\newblock


\bibitem[\protect\citeauthoryear{Cacciarelli and Kulahci}{Cacciarelli and
  Kulahci}{2024}]%
        {cacciarelli-ActiveLearningData-2024}
\bibfield{author}{\bibinfo{person}{Davide Cacciarelli} {and}
  \bibinfo{person}{Murat Kulahci}.} \bibinfo{year}{2024}\natexlab{}.
\newblock \showarticletitle{Active Learning for Data Streams: A Survey}.
\newblock \bibinfo{journal}{\emph{Machine Learning}} \bibinfo{volume}{113},
  \bibinfo{number}{1} (\bibinfo{date}{Jan.} \bibinfo{year}{2024}),
  \bibinfo{pages}{185--239}.
\newblock
\showISSN{1573-0565}
\urldef\tempurl%
\url{https://doi.org/10.1007/s10994-023-06454-2}
\showDOI{\tempurl}


\bibitem[\protect\citeauthoryear{Capit, Da~Costa, Georgiou, Huard, Martin,
  Mounie, Neyron, and Richard}{Capit et~al\mbox{.}}{2005}]%
        {oar}
\bibfield{author}{\bibinfo{person}{N. Capit}, \bibinfo{person}{G. Da~Costa},
  \bibinfo{person}{Y. Georgiou}, \bibinfo{person}{G. Huard},
  \bibinfo{person}{C. Martin}, \bibinfo{person}{G. Mounie}, \bibinfo{person}{P.
  Neyron}, {and} \bibinfo{person}{O. Richard}.}
  \bibinfo{year}{2005}\natexlab{}.
\newblock \showarticletitle{A batch scheduler with high level components}. In
  \bibinfo{booktitle}{\emph{CCGrid 2005. IEEE International Symposium on
  Cluster Computing and the Grid, 2005.}}, Vol.~\bibinfo{volume}{2}.
  \bibinfo{pages}{776--783 Vol. 2}.
\newblock
\urldef\tempurl%
\url{https://doi.org/10.1109/CCGRID.2005.1558641}
\showDOI{\tempurl}


\bibitem[\protect\citeauthoryear{Capp{\'e}, Guillin, Marin, and
  Robert}{Capp{\'e} et~al\mbox{.}}{2004}]%
        {cappe-PMC}
\bibfield{author}{\bibinfo{person}{Olivier Capp{\'e}}, \bibinfo{person}{Arnaud
  Guillin}, \bibinfo{person}{Jean-Michel Marin}, {and}
  \bibinfo{person}{Christian Robert}.} \bibinfo{year}{2004}\natexlab{}.
\newblock \showarticletitle{{Population Monte Carlo}}.
\newblock \bibinfo{journal}{\emph{{Journal of Computational and Graphical
  Statistics}}} \bibinfo{volume}{13}, \bibinfo{number}{4}
  (\bibinfo{year}{2004}), \bibinfo{pages}{907--929}.
\newblock
\urldef\tempurl%
\url{https://hal.science/hal-01337419}
\showURL{%
\tempurl}


\bibitem[\protect\citeauthoryear{{Cheng}, {Hao}, {Wang}, {Huang}, {Wu}, {Liu},
  {Zhao}, {Liu}, and {Su}}{{Cheng} et~al\mbox{.}}{2024}]%
        {referNO2024}
\bibfield{author}{\bibinfo{person}{Ze {Cheng}}, \bibinfo{person}{Zhongkai
  {Hao}}, \bibinfo{person}{Xiaoqiang {Wang}}, \bibinfo{person}{Jianing
  {Huang}}, \bibinfo{person}{Youjia {Wu}}, \bibinfo{person}{Xudan {Liu}},
  \bibinfo{person}{Yiru {Zhao}}, \bibinfo{person}{Songming {Liu}}, {and}
  \bibinfo{person}{Hang {Su}}.} \bibinfo{year}{2024}\natexlab{}.
\newblock \showarticletitle{{Reference Neural Operators: Learning the Smooth
  Dependence of Solutions of PDEs on Geometric Deformations}}.
\newblock \bibinfo{journal}{\emph{arXiv e-prints}}, Article
  \bibinfo{articleno}{arXiv:2405.17509} (\bibinfo{date}{May}
  \bibinfo{year}{2024}), \bibinfo{numpages}{arXiv:2405.17509}~pages.
\newblock
\urldef\tempurl%
\url{https://doi.org/10.48550/arXiv.2405.17509}
\showDOI{\tempurl}
\showeprint[arxiv]{2405.17509}~[cs.LG]


\bibitem[\protect\citeauthoryear{Cranmer, Brehmer, and Louppe}{Cranmer
  et~al\mbox{.}}{2020}]%
        {cranmer-FrontierSimulationBasedInference-2020}
\bibfield{author}{\bibinfo{person}{Kyle Cranmer}, \bibinfo{person}{Johann
  Brehmer}, {and} \bibinfo{person}{Gilles Louppe}.}
  \bibinfo{year}{2020}\natexlab{}.
\newblock \showarticletitle{The {{Frontier}} of {{Simulation-Based
  Inference}}}.
\newblock \bibinfo{journal}{\emph{Proceedings of the National Academy of
  Sciences}} \bibinfo{volume}{117}, \bibinfo{number}{48} (\bibinfo{date}{Dec.}
  \bibinfo{year}{2020}), \bibinfo{pages}{30055--30062}.
\newblock
\showISSN{0027-8424, 1091-6490}
\urldef\tempurl%
\url{https://doi.org/10.1073/pnas.1912789117}
\showDOI{\tempurl}


\bibitem[\protect\citeauthoryear{Daw, Bu, Wang, Perdikaris, and Karpatne}{Daw
  et~al\mbox{.}}{2023}]%
        {daw_mitigating_2023}
\bibfield{author}{\bibinfo{person}{Arka Daw}, \bibinfo{person}{Jie Bu},
  \bibinfo{person}{Sifan Wang}, \bibinfo{person}{Paris Perdikaris}, {and}
  \bibinfo{person}{Anuj Karpatne}.} \bibinfo{year}{2023}\natexlab{}.
\newblock \showarticletitle{Mitigating Propagation Failures in Physics-informed
  Neural Networks using Retain-Resample-Release (R3) Sampling}. In
  \bibinfo{booktitle}{\emph{Proceedings of the 40th International Conference on
  Machine Learning}}. \bibinfo{publisher}{{PMLR}}, \bibinfo{pages}{7264--7302}.
\newblock
\urldef\tempurl%
\url{https://proceedings.mlr.press/v202/daw23a.html}
\showURL{%
\tempurl}
\newblock
\shownote{{ISSN}: 2640-3498.}


\bibitem[\protect\citeauthoryear{Dolstra, De~Jonge, Visser,
  et~al\mbox{.}}{Dolstra et~al\mbox{.}}{2004}]%
        {dolstra2004nix}
\bibfield{author}{\bibinfo{person}{Eelco Dolstra}, \bibinfo{person}{Merijn
  De~Jonge}, \bibinfo{person}{Eelco Visser}, {et~al\mbox{.}}}
  \bibinfo{year}{2004}\natexlab{}.
\newblock \showarticletitle{Nix: A Safe and Policy-Free System for Software
  Deployment.}. In \bibinfo{booktitle}{\emph{LISA}}, Vol.~\bibinfo{volume}{4}.
  \bibinfo{pages}{79--92}.
\newblock


\bibitem[\protect\citeauthoryear{Dymchenko and Raffin}{Dymchenko and
  Raffin}{2023}]%
        {myself}
\bibfield{author}{\bibinfo{person}{Sofya Dymchenko} {and}
  \bibinfo{person}{Bruno Raffin}.} \bibinfo{year}{2023}\natexlab{}.
\newblock \showarticletitle{Loss-driven sampling within hard-to-learn areas for
  simulation-based neural network training}. In
  \bibinfo{booktitle}{\emph{{MLPS} 2023 - {Machine} {Learning} and the
  {Physical} {Sciences} {Workshop} at {NeurIPS} 2023 - 37th conference on
  {Neural} {Information} {Processing} {Systems}}}. \bibinfo{address}{New
  Orleans, United States}, \bibinfo{pages}{1--5}.
\newblock
\urldef\tempurl%
\url{https://hal.science/hal-04305233}
\showURL{%
\tempurl}


\bibitem[\protect\citeauthoryear{Elvira and Chouzenoux}{Elvira and
  Chouzenoux}{2022}]%
        {elvira-OptimizedPopulationMonte-2022}
\bibfield{author}{\bibinfo{person}{V{\'i}ctor Elvira} {and}
  \bibinfo{person}{{\'E}milie Chouzenoux}.} \bibinfo{year}{2022}\natexlab{}.
\newblock \showarticletitle{Optimized {{Population Monte Carlo}}}.
\newblock \bibinfo{journal}{\emph{IEEE Transactions on Signal Processing}}
  \bibinfo{volume}{70} (\bibinfo{year}{2022}), \bibinfo{pages}{2489--2501}.
\newblock
\showISSN{1053-587X, 1941-0476}
\urldef\tempurl%
\url{https://doi.org/10.1109/TSP.2022.3172619}
\showDOI{\tempurl}
\showeprint[arxiv]{2204.06891}~[stat]


\bibitem[\protect\citeauthoryear{Elvira, Chouzenoux, Akyildiz, and
  Martino}{Elvira et~al\mbox{.}}{2023}]%
        {elvira-GradientbasedAdaptiveImportance-2023}
\bibfield{author}{\bibinfo{person}{V{\'i}ctor Elvira},
  \bibinfo{person}{{\'E}milie Chouzenoux}, \bibinfo{person}{{\"O}mer~Deniz
  Akyildiz}, {and} \bibinfo{person}{Luca Martino}.}
  \bibinfo{year}{2023}\natexlab{}.
\newblock \showarticletitle{Gradient-Based Adaptive Importance Samplers}.
\newblock \bibinfo{journal}{\emph{Journal of the Franklin Institute}}
  \bibinfo{volume}{360}, \bibinfo{number}{13} (\bibinfo{date}{Sept.}
  \bibinfo{year}{2023}), \bibinfo{pages}{9490--9514}.
\newblock
\showISSN{0016-0032}
\urldef\tempurl%
\url{https://doi.org/10.1016/j.jfranklin.2023.06.041}
\showDOI{\tempurl}


\bibitem[\protect\citeauthoryear{Fayyaz, Aghazadeh, Modarressi, Pilehvar,
  Yaghoobzadeh, and Kahou}{Fayyaz et~al\mbox{.}}{2022}]%
        {fayyaz_bert_2022}
\bibfield{author}{\bibinfo{person}{Mohsen Fayyaz}, \bibinfo{person}{Ehsan
  Aghazadeh}, \bibinfo{person}{Ali Modarressi}, \bibinfo{person}{Mohammad~Taher
  Pilehvar}, \bibinfo{person}{Yadollah Yaghoobzadeh}, {and}
  \bibinfo{person}{Samira~Ebrahimi Kahou}.} \bibinfo{year}{2022}\natexlab{}.
\newblock \showarticletitle{{BERT} on a Data Diet: Finding Important Examples
  by Gradient-Based Pruning}. In \bibinfo{booktitle}{\emph{{NeurIPS}}}.
\newblock


\bibitem[\protect\citeauthoryear{Herde, Raoni{\'c}, Rohner, K{\"a}ppeli,
  Molinaro, {de B{\'e}zenac}, and Mishra}{Herde et~al\mbox{.}}{2024}]%
        {herde-PoseidonEfficientFoundation-2024}
\bibfield{author}{\bibinfo{person}{Maximilian Herde}, \bibinfo{person}{Bogdan
  Raoni{\'c}}, \bibinfo{person}{Tobias Rohner}, \bibinfo{person}{Roger
  K{\"a}ppeli}, \bibinfo{person}{Roberto Molinaro}, \bibinfo{person}{Emmanuel
  {de B{\'e}zenac}}, {and} \bibinfo{person}{Siddhartha Mishra}.}
  \bibinfo{year}{2024}\natexlab{}.
\newblock \bibinfo{title}{Poseidon: {{Efficient Foundation Models}} for
  {{PDEs}}}.
\newblock
\newblock
\urldef\tempurl%
\url{https://doi.org/10.48550/arXiv.2405.19101}
\showDOI{\tempurl}
\showeprint[arxiv]{2405.19101}~[cs]


\bibitem[\protect\citeauthoryear{K and Søgaard}{K and Søgaard}{2021}]%
        {k_revisiting_2021}
\bibfield{author}{\bibinfo{person}{Karthikeyan K} {and} \bibinfo{person}{Anders
  Søgaard}.} \bibinfo{year}{2021}\natexlab{}.
\newblock \showarticletitle{Revisiting Methods for Finding Influential
  Examples}.
\newblock  (\bibinfo{year}{2021}).
\newblock
\urldef\tempurl%
\url{https://doi.org/10.48550/arxiv.2111.04683}
\showDOI{\tempurl}
\showeprint[arxiv]{2111.04683}


\bibitem[\protect\citeauthoryear{Katharopoulos and Fleuret}{Katharopoulos and
  Fleuret}{2019}]%
        {katharopoulos_not_2019}
\bibfield{author}{\bibinfo{person}{Angelos Katharopoulos} {and}
  \bibinfo{person}{François Fleuret}.} \bibinfo{year}{2019}\natexlab{}.
\newblock \bibinfo{title}{Not All Samples Are Created Equal: Deep Learning with
  Importance Sampling}.
\newblock
\newblock
\urldef\tempurl%
\url{https://doi.org/10.48550/arXiv.1803.00942}
\showDOI{\tempurl}
\showeprint[arxiv]{1803.00942 [cs]}


\bibitem[\protect\citeauthoryear{Killamsetty, Abhishek, Ramakrishnan,
  Evfimievski, Popa, and Iyer}{Killamsetty et~al\mbox{.}}{2022}]%
        {killamsetty_automata_2022}
\bibfield{author}{\bibinfo{person}{Krishnateja Killamsetty},
  \bibinfo{person}{Guttu~Sai Abhishek}, \bibinfo{person}{Ganesh Ramakrishnan},
  \bibinfo{person}{Alexandre~V Evfimievski}, \bibinfo{person}{Lucian Popa},
  {and} \bibinfo{person}{Rishabh Iyer}.} \bibinfo{year}{2022}\natexlab{}.
\newblock \showarticletitle{{AUTOMATA} : Gradient Based Data Subset Selection
  for Compute-Efficient Hyper-parameter Tuning}. In
  \bibinfo{booktitle}{\emph{Advances in Neural Information Processing
  Systems}}.
\newblock


\bibitem[\protect\citeauthoryear{Kochkov, Yuval, Langmore, Norgaard, Smith,
  Mooers, Kl{\"o}wer, Lottes, Rasp, D{\"u}ben, Hatfield, Battaglia,
  {Sanchez-Gonzalez}, Willson, Brenner, and Hoyer}{Kochkov
  et~al\mbox{.}}{2024}]%
        {kochkov-NeuralGeneralCirculation-2024a}
\bibfield{author}{\bibinfo{person}{Dmitrii Kochkov}, \bibinfo{person}{Janni
  Yuval}, \bibinfo{person}{Ian Langmore}, \bibinfo{person}{Peter Norgaard},
  \bibinfo{person}{Jamie Smith}, \bibinfo{person}{Griffin Mooers},
  \bibinfo{person}{Milan Kl{\"o}wer}, \bibinfo{person}{James Lottes},
  \bibinfo{person}{Stephan Rasp}, \bibinfo{person}{Peter D{\"u}ben},
  \bibinfo{person}{Sam Hatfield}, \bibinfo{person}{Peter Battaglia},
  \bibinfo{person}{Alvaro {Sanchez-Gonzalez}}, \bibinfo{person}{Matthew
  Willson}, \bibinfo{person}{Michael~P. Brenner}, {and}
  \bibinfo{person}{Stephan Hoyer}.} \bibinfo{year}{2024}\natexlab{}.
\newblock \showarticletitle{Neural General Circulation Models for Weather and
  Climate}.
\newblock \bibinfo{journal}{\emph{Nature}} (\bibinfo{date}{July}
  \bibinfo{year}{2024}).
\newblock
\showISSN{1476-4687}
\urldef\tempurl%
\url{https://doi.org/10.1038/s41586-024-07744-y}
\showDOI{\tempurl}


\bibitem[\protect\citeauthoryear{{Kohl}, {Chen}, and {Thuerey}}{{Kohl}
  et~al\mbox{.}}{2023}]%
        {kohl-DiffFlow-2023}
\bibfield{author}{\bibinfo{person}{Georg {Kohl}}, \bibinfo{person}{Li-Wei
  {Chen}}, {and} \bibinfo{person}{Nils {Thuerey}}.}
  \bibinfo{year}{2023}\natexlab{}.
\newblock \showarticletitle{{Benchmarking Autoregressive Conditional Diffusion
  Models for Turbulent Flow Simulation}}.
\newblock \bibinfo{journal}{\emph{arXiv e-prints}}, Article
  \bibinfo{articleno}{arXiv:2309.01745} (\bibinfo{date}{Sept.}
  \bibinfo{year}{2023}), \bibinfo{numpages}{arXiv:2309.01745}~pages.
\newblock
\urldef\tempurl%
\url{https://doi.org/10.48550/arXiv.2309.01745}
\showDOI{\tempurl}
\showeprint[arxiv]{2309.01745}~[cs.LG]


\bibitem[\protect\citeauthoryear{Kye, Choi, and Chang}{Kye
  et~al\mbox{.}}{2022}]%
        {kye_tidal_2022}
\bibfield{author}{\bibinfo{person}{Seong~Min Kye}, \bibinfo{person}{Kwanghee
  Choi}, {and} \bibinfo{person}{Buru Chang}.} \bibinfo{year}{2022}\natexlab{}.
\newblock \showarticletitle{{TiDAL}: Learning Training Dynamics for Active
  Learning}.
\newblock  (\bibinfo{year}{2022}).
\newblock
\urldef\tempurl%
\url{https://doi.org/10.48550/ARXIV.2210.06788}
\showDOI{\tempurl}
\newblock
\shownote{Publisher: {arXiv} Version Number: 1.}


\bibitem[\protect\citeauthoryear{Lau, Hemachandra, Ng, and Low}{Lau
  et~al\mbox{.}}{2023}]%
        {lau-PINNACLEPINNAdaptive-2023}
\bibfield{author}{\bibinfo{person}{Gregory Kang~Ruey Lau},
  \bibinfo{person}{Apivich Hemachandra}, \bibinfo{person}{See-Kiong Ng}, {and}
  \bibinfo{person}{Bryan Kian~Hsiang Low}.} \bibinfo{year}{2023}\natexlab{}.
\newblock \showarticletitle{{{PINNACLE}}: {{PINN Adaptive ColLocation}} and
  {{Experimental}} Points Selection}. In \bibinfo{booktitle}{\emph{The
  {{Twelfth International Conference}} on {{Learning Representations}}}}.
\newblock


\bibitem[\protect\citeauthoryear{Lavin, Zenil, Paige, Krakauer, Gottschlich,
  Mattson, Anandkumar, Choudry, Rocki, Baydin, Prunkl, Paige, Isayev, Peterson,
  {McMahon}, Macke, Cranmer, Zhang, Wainwright, Hanuka, Veloso, Assefa, Zheng,
  and Pfeffer}{Lavin et~al\mbox{.}}{2021}]%
        {lavin_simulation_2021}
\bibfield{author}{\bibinfo{person}{Alexander Lavin}, \bibinfo{person}{Hector
  Zenil}, \bibinfo{person}{Brooks Paige}, \bibinfo{person}{David Krakauer},
  \bibinfo{person}{Justin Gottschlich}, \bibinfo{person}{Tim Mattson},
  \bibinfo{person}{Anima Anandkumar}, \bibinfo{person}{Sanjay Choudry},
  \bibinfo{person}{Kamil Rocki}, \bibinfo{person}{Atılım~Güneş Baydin},
  \bibinfo{person}{Carina Prunkl}, \bibinfo{person}{Brooks Paige},
  \bibinfo{person}{Olexandr Isayev}, \bibinfo{person}{Erik Peterson},
  \bibinfo{person}{Peter~L. {McMahon}}, \bibinfo{person}{Jakob Macke},
  \bibinfo{person}{Kyle Cranmer}, \bibinfo{person}{Jiaxin Zhang},
  \bibinfo{person}{Haruko Wainwright}, \bibinfo{person}{Adi Hanuka},
  \bibinfo{person}{Manuela Veloso}, \bibinfo{person}{Samuel Assefa},
  \bibinfo{person}{Stephan Zheng}, {and} \bibinfo{person}{Avi Pfeffer}.}
  \bibinfo{year}{2021}\natexlab{}.
\newblock \showarticletitle{Simulation Intelligence: Towards a New Generation
  of Scientific Methods}.
\newblock  (\bibinfo{year}{2021}).
\newblock
\showeprint[arxiv]{2112.03235}
\urldef\tempurl%
\url{http://arxiv.org/abs/2112.03235}
\showURL{%
\tempurl}


\bibitem[\protect\citeauthoryear{Li, Kovachki, Azizzadenesheli, Liu,
  Bhattacharya, Stuart, and Anandkumar}{Li et~al\mbox{.}}{2021}]%
        {li2020fourier}
\bibfield{author}{\bibinfo{person}{Zongyi Li},
  \bibinfo{person}{Nikola~Borislavov Kovachki}, \bibinfo{person}{Kamyar
  Azizzadenesheli}, \bibinfo{person}{Burigede Liu}, \bibinfo{person}{Kaushik
  Bhattacharya}, \bibinfo{person}{Andrew~M. Stuart}, {and}
  \bibinfo{person}{Anima Anandkumar}.} \bibinfo{year}{2021}\natexlab{}.
\newblock \showarticletitle{Fourier Neural Operator for Parametric Partial
  Differential Equations}. In \bibinfo{booktitle}{\emph{9th International
  Conference on Learning Representations, {ICLR} 2021, Virtual Event, Austria,
  May 3-7, 2021}}. \bibinfo{publisher}{OpenReview.net}.
\newblock
\urldef\tempurl%
\url{https://openreview.net/forum?id=c8P9NQVtmnO}
\showURL{%
\tempurl}


\bibitem[\protect\citeauthoryear{Meyer, Schouler, Caulk, Rib{\'e}s, and
  Raffin}{Meyer et~al\mbox{.}}{2023a}]%
        {meyer-SC23}
\bibfield{author}{\bibinfo{person}{Lucas Meyer}, \bibinfo{person}{Marc
  Schouler}, \bibinfo{person}{Robert~Alexander Caulk},
  \bibinfo{person}{Alejandro Rib{\'e}s}, {and} \bibinfo{person}{Bruno Raffin}.}
  \bibinfo{year}{2023}\natexlab{a}.
\newblock \showarticletitle{{High Throughput Training of Deep Surrogates from
  Large Ensemble Runs}}. In \bibinfo{booktitle}{\emph{{SC 2023 - The
  International Conference for High Performance Computing, Networking, Storage,
  and Analysis}}}. \bibinfo{publisher}{{ACM}}, \bibinfo{address}{Denver, CO,
  United States}, \bibinfo{pages}{1--14}.
\newblock
\urldef\tempurl%
\url{https://doi.org/10.1145/3581784.3607083}
\showDOI{\tempurl}


\bibitem[\protect\citeauthoryear{Meyer, Schouler, Caulk, Rib{\'e}s, and
  Raffin}{Meyer et~al\mbox{.}}{2023b}]%
        {meyer-ICML23}
\bibfield{author}{\bibinfo{person}{Lucas Meyer}, \bibinfo{person}{Marc
  Schouler}, \bibinfo{person}{Robert~Alexander Caulk},
  \bibinfo{person}{Alejandro Rib{\'e}s}, {and} \bibinfo{person}{Bruno Raffin}.}
  \bibinfo{year}{2023}\natexlab{b}.
\newblock \showarticletitle{{Training Deep Surrogate Models with Large Scale
  Online Learning}}. In \bibinfo{booktitle}{\emph{{ICML 2023 - International
  Conference on Machine Learning}}}. \bibinfo{pages}{1--17}.
\newblock
\urldef\tempurl%
\url{https://hal.science/hal-04102400}
\showURL{%
\tempurl}


\bibitem[\protect\citeauthoryear{Musekamp, Kalimuthu, Holzm{\"u}ller, Takamoto,
  and Niepert}{Musekamp et~al\mbox{.}}{2024}]%
        {musekamp-ActiveLearningNeural-2024}
\bibfield{author}{\bibinfo{person}{Daniel Musekamp}, \bibinfo{person}{Marimuthu
  Kalimuthu}, \bibinfo{person}{David Holzm{\"u}ller}, \bibinfo{person}{Makoto
  Takamoto}, {and} \bibinfo{person}{Mathias Niepert}.}
  \bibinfo{year}{2024}\natexlab{}.
\newblock \bibinfo{title}{Active {{Learning}} for {{Neural PDE Solvers}}}.
\newblock
\newblock
\showeprint[arxiv]{2408.01536}~[cs]


\bibitem[\protect\citeauthoryear{Mölder, Jablonski, Letcher, Hall,
  Tomkins-Tinch, Sochat, Forster, Lee, Twardziok, Kanitz, Wilm, Holtgrewe,
  Rahmann, Nahnsen, and Köster}{Mölder et~al\mbox{.}}{2021}]%
        {snakemake}
\bibfield{author}{\bibinfo{person}{F Mölder}, \bibinfo{person}{KP Jablonski},
  \bibinfo{person}{B Letcher}, \bibinfo{person}{MB Hall}, \bibinfo{person}{CH
  Tomkins-Tinch}, \bibinfo{person}{V Sochat}, \bibinfo{person}{J Forster},
  \bibinfo{person}{S Lee}, \bibinfo{person}{SO Twardziok}, \bibinfo{person}{A
  Kanitz}, \bibinfo{person}{A Wilm}, \bibinfo{person}{M Holtgrewe},
  \bibinfo{person}{S Rahmann}, \bibinfo{person}{S Nahnsen}, {and}
  \bibinfo{person}{J Köster}.} \bibinfo{year}{2021}\natexlab{}.
\newblock \showarticletitle{Sustainable data analysis with Snakemake [version
  2; peer review: 2 approved]}.
\newblock \bibinfo{journal}{\emph{F1000Research}} \bibinfo{volume}{10},
  \bibinfo{number}{33} (\bibinfo{year}{2021}).
\newblock
\urldef\tempurl%
\url{https://doi.org/10.12688/f1000research.29032.2}
\showDOI{\tempurl}


\bibitem[\protect\citeauthoryear{Nabian, Gladstone, and Meidani}{Nabian
  et~al\mbox{.}}{2021}]%
        {nabian_efficient_2021}
\bibfield{author}{\bibinfo{person}{Mohammad~Amin Nabian},
  \bibinfo{person}{Rini~Jasmine Gladstone}, {and} \bibinfo{person}{Hadi
  Meidani}.} \bibinfo{year}{2021}\natexlab{}.
\newblock \showarticletitle{Efficient training of physics-informed neural
  networks via importance sampling}.
\newblock  \bibinfo{volume}{36}, \bibinfo{number}{8} (\bibinfo{year}{2021}),
  \bibinfo{pages}{962--977}.
\newblock
\showISSN{1093-9687, 1467-8667}
\urldef\tempurl%
\url{https://doi.org/10.1111/mice.12685}
\showDOI{\tempurl}
\showeprint[arxiv]{2104.12325 [cs, math]}


\bibitem[\protect\citeauthoryear{Nguyen, Brandstetter, Kapoor, Gupta, and
  Grover}{Nguyen et~al\mbox{.}}{2023}]%
        {nguyen-ClimaXFoundationModel-2023}
\bibfield{author}{\bibinfo{person}{Tung Nguyen}, \bibinfo{person}{Johannes
  Brandstetter}, \bibinfo{person}{Ashish Kapoor}, \bibinfo{person}{Jayesh~K
  Gupta}, {and} \bibinfo{person}{Aditya Grover}.}
  \bibinfo{year}{2023}\natexlab{}.
\newblock \showarticletitle{{C}lima{X}: A foundation model for weather and
  climate}. In \bibinfo{booktitle}{\emph{Proceedings of the 40th International
  Conference on Machine Learning}} \emph{(\bibinfo{series}{Proceedings of
  Machine Learning Research}, Vol.~\bibinfo{volume}{202})},
  \bibfield{editor}{\bibinfo{person}{Andreas Krause}, \bibinfo{person}{Emma
  Brunskill}, \bibinfo{person}{Kyunghyun Cho}, \bibinfo{person}{Barbara
  Engelhardt}, \bibinfo{person}{Sivan Sabato}, {and} \bibinfo{person}{Jonathan
  Scarlett}} (Eds.). \bibinfo{publisher}{PMLR}, \bibinfo{pages}{25904--25938}.
\newblock
\urldef\tempurl%
\url{https://proceedings.mlr.press/v202/nguyen23a.html}
\showURL{%
\tempurl}


\bibitem[\protect\citeauthoryear{Pal, Valkanas, and Coates}{Pal
  et~al\mbox{.}}{2023}]%
        {pal_population_2023}
\bibfield{author}{\bibinfo{person}{Soumyasundar Pal}, \bibinfo{person}{Antonios
  Valkanas}, {and} \bibinfo{person}{Mark Coates}.}
  \bibinfo{year}{2023}\natexlab{}.
\newblock \bibinfo{title}{Population {Monte} {Carlo} with {Normalizing}
  {Flow}}.
\newblock
\newblock
\urldef\tempurl%
\url{https://doi.org/10.48550/arXiv.2312.03857}
\showDOI{\tempurl}
\newblock
\shownote{arXiv:2312.03857 [stat].}


\bibitem[\protect\citeauthoryear{Pfaff, Fortunato, Sanchez{-}Gonzalez, and
  Battaglia}{Pfaff et~al\mbox{.}}{2021}]%
        {pfaff2020learning}
\bibfield{author}{\bibinfo{person}{Tobias Pfaff}, \bibinfo{person}{Meire
  Fortunato}, \bibinfo{person}{Alvaro Sanchez{-}Gonzalez}, {and}
  \bibinfo{person}{Peter~W. Battaglia}.} \bibinfo{year}{2021}\natexlab{}.
\newblock \showarticletitle{Learning Mesh-Based Simulation with Graph
  Networks}. In \bibinfo{booktitle}{\emph{9th International Conference on
  Learning Representations, {ICLR} 2021, Virtual Event, Austria, May 3-7,
  2021}}. \bibinfo{publisher}{OpenReview.net}.
\newblock
\urldef\tempurl%
\url{https://openreview.net/forum?id=roNqYL0\_XP}
\showURL{%
\tempurl}


\bibitem[\protect\citeauthoryear{Raissi, Perdikaris, and Karniadakis}{Raissi
  et~al\mbox{.}}{2019}]%
        {RAISSI2019686}
\bibfield{author}{\bibinfo{person}{M. Raissi}, \bibinfo{person}{P. Perdikaris},
  {and} \bibinfo{person}{G.E. Karniadakis}.} \bibinfo{year}{2019}\natexlab{}.
\newblock \showarticletitle{Physics-informed neural networks: A deep learning
  framework for solving forward and inverse problems involving nonlinear
  partial differential equations}.
\newblock \bibinfo{journal}{\emph{J. Comput. Phys.}}  \bibinfo{volume}{378}
  (\bibinfo{year}{2019}), \bibinfo{pages}{686--707}.
\newblock
\showISSN{0021-9991}
\urldef\tempurl%
\url{https://doi.org/10.1016/j.jcp.2018.10.045}
\showDOI{\tempurl}


\bibitem[\protect\citeauthoryear{{Ren}, {Xiao}, {Chang}, {Huang}, {Li},
  {Gupta}, {Chen}, and {Wang}}{{Ren} et~al\mbox{.}}{2020}]%
        {actdl2020}
\bibfield{author}{\bibinfo{person}{Pengzhen {Ren}}, \bibinfo{person}{Yun
  {Xiao}}, \bibinfo{person}{Xiaojun {Chang}}, \bibinfo{person}{Po-Yao {Huang}},
  \bibinfo{person}{Zhihui {Li}}, \bibinfo{person}{Brij~B. {Gupta}},
  \bibinfo{person}{Xiaojiang {Chen}}, {and} \bibinfo{person}{Xin {Wang}}.}
  \bibinfo{year}{2020}\natexlab{}.
\newblock \showarticletitle{{A Survey of Deep Active Learning}}.
\newblock \bibinfo{journal}{\emph{arXiv e-prints}}, Article
  \bibinfo{articleno}{arXiv:2009.00236} (\bibinfo{date}{Aug.}
  \bibinfo{year}{2020}), \bibinfo{numpages}{arXiv:2009.00236}~pages.
\newblock
\urldef\tempurl%
\url{https://doi.org/10.48550/arXiv.2009.00236}
\showDOI{\tempurl}
\showeprint[arxiv]{2009.00236}~[cs.LG]


\bibitem[\protect\citeauthoryear{Schouler, Caulk, Meyer, Terraz, Conrads,
  Friedemann, Agarwal, Baldonado, Pogodziński, Sekuła, Ribes, and
  Raffin}{Schouler et~al\mbox{.}}{2023}]%
        {Schouler-JOSS23}
\bibfield{author}{\bibinfo{person}{Marc Schouler},
  \bibinfo{person}{Robert~Alexander Caulk}, \bibinfo{person}{Lucas Meyer},
  \bibinfo{person}{Théophile Terraz}, \bibinfo{person}{Christoph Conrads},
  \bibinfo{person}{Sebastian Friedemann}, \bibinfo{person}{Achal Agarwal},
  \bibinfo{person}{Juan~Manuel Baldonado}, \bibinfo{person}{Bartłomiej
  Pogodziński}, \bibinfo{person}{Anna Sekuła}, \bibinfo{person}{Alejandro
  Ribes}, {and} \bibinfo{person}{Bruno Raffin}.}
  \bibinfo{year}{2023}\natexlab{}.
\newblock \showarticletitle{Melissa: coordinating large-scale ensemble runs for
  deep learning and sensitivity analyses}.
\newblock \bibinfo{journal}{\emph{Journal of Open Source Software}}
  \bibinfo{volume}{8}, \bibinfo{number}{86} (\bibinfo{year}{2023}),
  \bibinfo{pages}{5291}.
\newblock
\urldef\tempurl%
\url{https://doi.org/10.21105/joss.05291}
\showDOI{\tempurl}


\bibitem[\protect\citeauthoryear{Takezoe, Liu, Mao, Chen, Feng, Zhang, and
  Wang}{Takezoe et~al\mbox{.}}{2023}]%
        {Takezoe-DeepActiveLearning-2023}
\bibfield{author}{\bibinfo{person}{Rinyoichi Takezoe}, \bibinfo{person}{Xu
  Liu}, \bibinfo{person}{Shunan Mao}, \bibinfo{person}{Marco~Tianyu Chen},
  \bibinfo{person}{Zhanpeng Feng}, \bibinfo{person}{Shiliang Zhang}, {and}
  \bibinfo{person}{Xiaoyu Wang}.} \bibinfo{year}{2023}\natexlab{}.
\newblock \showarticletitle{Deep Active Learning for Computer Vision: Past and
  Future}.
\newblock \bibinfo{journal}{\emph{APSIPA Transactions on Signal and Information
  Processing}} \bibinfo{volume}{12}, \bibinfo{number}{1}
  (\bibinfo{year}{2023}), \bibinfo{pages}{--}.
\newblock
\urldef\tempurl%
\url{https://doi.org/10.1561/116.00000057}
\showDOI{\tempurl}


\bibitem[\protect\citeauthoryear{Wang, Huang, Wu, Tong, Margenot, and He}{Wang
  et~al\mbox{.}}{2022}]%
        {wang_deep_2022}
\bibfield{author}{\bibinfo{person}{Haonan Wang}, \bibinfo{person}{Wei Huang},
  \bibinfo{person}{Ziwei Wu}, \bibinfo{person}{Hanghang Tong},
  \bibinfo{person}{Andrew~J. Margenot}, {and} \bibinfo{person}{Jingrui He}.}
  \bibinfo{year}{2022}\natexlab{}.
\newblock \showarticletitle{Deep Active Learning by Leveraging Training
  Dynamics}.
\newblock
\urldef\tempurl%
\url{https://openreview.net/forum?id=aJ5xc1QB7EX}
\showURL{%
\tempurl}


\bibitem[\protect\citeauthoryear{Wang, Sankaran, Wang, and Perdikaris}{Wang
  et~al\mbox{.}}{2023}]%
        {wang_experts_2023}
\bibfield{author}{\bibinfo{person}{Sifan Wang}, \bibinfo{person}{Shyam
  Sankaran}, \bibinfo{person}{Hanwen Wang}, {and} \bibinfo{person}{Paris
  Perdikaris}.} \bibinfo{year}{2023}\natexlab{}.
\newblock \bibinfo{title}{An {Expert}'s {Guide} to {Training}
  {Physics}-informed {Neural} {Networks}}.
\newblock
\newblock
\urldef\tempurl%
\url{https://doi.org/10.48550/arXiv.2308.08468}
\showDOI{\tempurl}
\newblock
\shownote{arXiv:2308.08468 [physics].}


\bibitem[\protect\citeauthoryear{Ward, Sivaraman, Pauloski, Babuji, Chard,
  Dandu, Redfern, Assary, Chard, Curtiss, Thakur, and Foster}{Ward
  et~al\mbox{.}}{2021}]%
        {ward_colmena_2021}
\bibfield{author}{\bibinfo{person}{Logan Ward}, \bibinfo{person}{Ganesh
  Sivaraman}, \bibinfo{person}{J.~Gregory Pauloski}, \bibinfo{person}{Yadu
  Babuji}, \bibinfo{person}{Ryan Chard}, \bibinfo{person}{Naveen Dandu},
  \bibinfo{person}{Paul~C. Redfern}, \bibinfo{person}{Rajeev~S. Assary},
  \bibinfo{person}{Kyle Chard}, \bibinfo{person}{Larry~A. Curtiss},
  \bibinfo{person}{Rajeev Thakur}, {and} \bibinfo{person}{Ian Foster}.}
  \bibinfo{year}{2021}\natexlab{}.
\newblock \showarticletitle{Colmena: Scalable Machine-Learning-Based Steering
  of Ensemble Simulations for High Performance Computing}.
\newblock  (\bibinfo{year}{2021}), \bibinfo{pages}{9--20}.
\newblock
\urldef\tempurl%
\url{https://doi.org/10.1109/MLHPC54614.2021.00007}
\showDOI{\tempurl}


\bibitem[\protect\citeauthoryear{Wu, Zhu, Tan, Kartha, and Lu}{Wu
  et~al\mbox{.}}{2022}]%
        {wu_comprehensive_2022}
\bibfield{author}{\bibinfo{person}{Chenxi Wu}, \bibinfo{person}{Min Zhu},
  \bibinfo{person}{Qinyang Tan}, \bibinfo{person}{Yadhu Kartha}, {and}
  \bibinfo{person}{Lu Lu}.} \bibinfo{year}{2022}\natexlab{}.
\newblock \bibinfo{title}{A comprehensive study of non-adaptive and
  residual-based adaptive sampling for physics-informed neural networks}.
\newblock
\newblock
\urldef\tempurl%
\url{https://doi.org/10.48550/arXiv.2207.10289}
\showDOI{\tempurl}
\showeprint[arxiv]{2207.10289 [physics]}


\bibitem[\protect\citeauthoryear{Yang, Qiu, and Fu}{Yang et~al\mbox{.}}{2022}]%
        {yang_dmis_2022}
\bibfield{author}{\bibinfo{person}{Zijiang Yang}, \bibinfo{person}{Zhongwei
  Qiu}, {and} \bibinfo{person}{Dongmei Fu}.} \bibinfo{year}{2022}\natexlab{}.
\newblock \bibinfo{title}{{DMIS}: Dynamic Mesh-based Importance Sampling for
  Training Physics-Informed Neural Networks}.
\newblock
\newblock
\urldef\tempurl%
\url{https://doi.org/10.48550/arXiv.2211.13944}
\showDOI{\tempurl}
\showeprint[arxiv]{2211.13944 [cs, math]}


\end{thebibliography}

\appendix

\section{Melissa DL Architecture}\label{app:melissa-details}
\begin{figure}[H]
	\centering
	\includegraphics[width=0.4\textwidth]{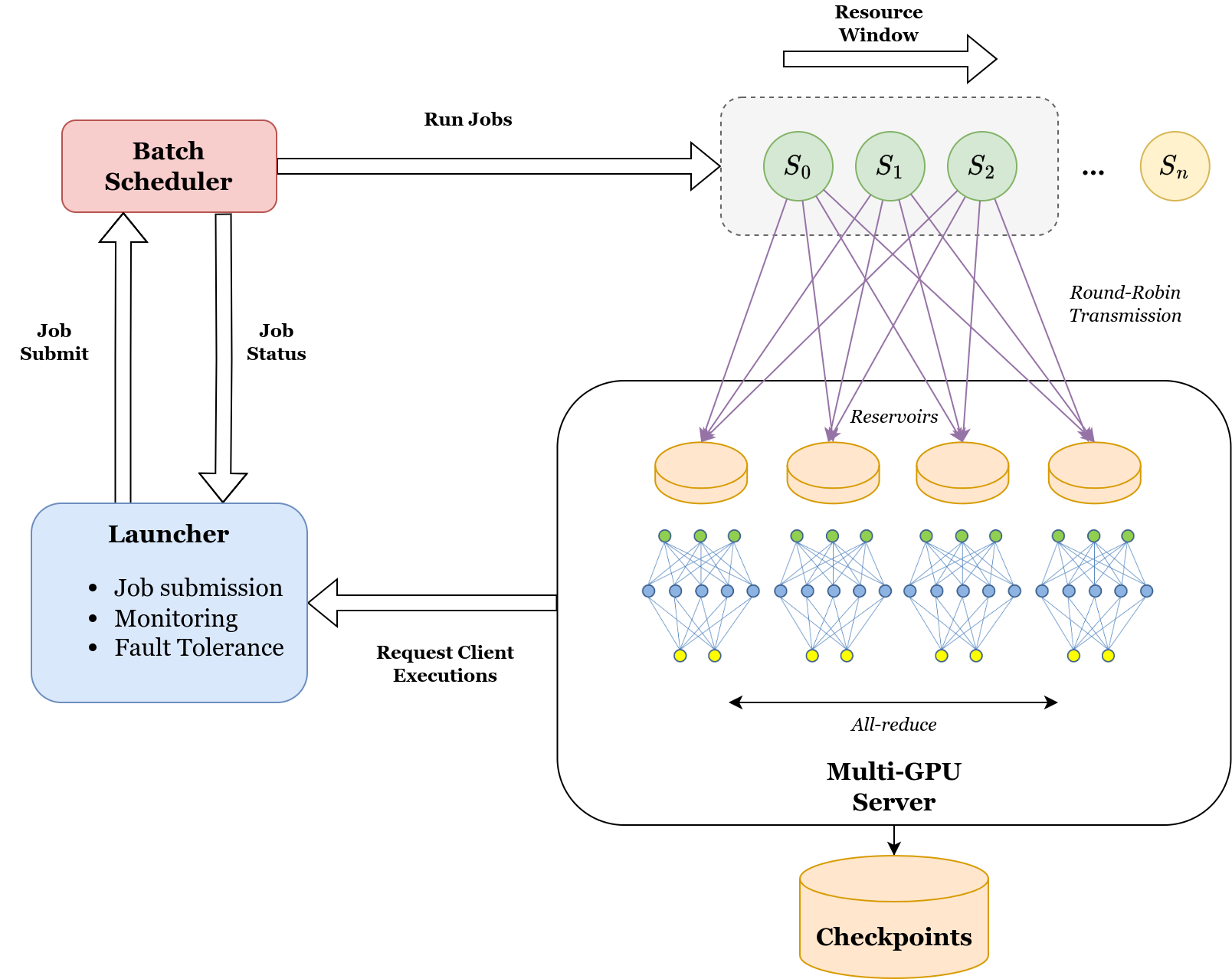}
	\caption{The Melissa framework architecture consisting of \textit{launcher} (orchestrates the process through the cluster's batch scheduler), \textit{clients jobs} (where simulations are executed), and  a \textit{server} (trains NN and manages  \textit{reservoirs}).
	}
	\label{fig:melissa}
\end{figure}
The \textit{launcher} controls the workflow by interacting with the  cluster's batch scheduler. It initiates the \textit{server}, a Python-based code using PyTorch for multi-GPU training.
The server manages the training process and defines the simulation instances, known as \textit{clients}. The server then sends a request to the launcher for submitting clients based on allocated resources. Once a client starts, it connects dynamically to the server.

The server maintains a memory buffer called the \textit{reservoir}, which  goal is to reduce training bias and avoid GPU starvation.
Newly received data from the clients are stored in the reservoir, replacing older data randomly. If all reservoir samples are new, client executions are  paused temporarily.
The server asynchronously creates random batches from the reservoir for NN training, allowing each reservoir sample to be reused multiple times. See~\cite{meyer-SC23}
for a detailed description of the reservoir algorithm. 

\section{General setup details}
\subsection{2D HeatPDE with Melissa}\label{app:general-setup-heatpde}
The experiments consider \textit{the classical heat equation (HeatPDE)} on a 2D rectangular domain:
\begin{align}
	&\frac{\partial u(x,t)}{\partial t} = \alpha \frac{\partial^2 u(x,t) }{\partial x^2}\\
	&u((x_1=0, x_2), t) = T_{1}, u((x_1=L, x_2), t) = T_{2}\\
	&u((x_1, x_2=0), t) = T_{3}, u((x_1, x_2=L), t) = T_{4}\\
	&u(x, t=0) = T_{0}
\end{align}

where $u(x,t)$ is the field temperature, $\alpha$ is the thermal diffusivity and $[T_0, T_1, T_2, T_3, T_4]$ are the initial and 4 boundary temperatures. The solution is approximated with an in-house solver that implements a finite difference method with an implicit Euler scheme. The temperature field is discretized on a $M\times M$ Cartesian grid and generated for $T=100$ time steps representing $\triangle t=0.01$ seconds each. In this study, the thermal diffusivity is fixed to $\alpha=1m^2.s^{-1}$, and changing this parameter is left for future work. The temperature parameters are the solver input parameters $\Lambda = \mathbb R^5$ that we tend to sample, which values we bound to interval $[100,500]$~K. \hide{For the validation set, we use 200 full-trajectory simulations with parameters generated with quasi-uniform Halton Sequence distribution that ensures good coverage of $\Lambda$.}

The surrogate is trained to directly predict the temperature field. As the initial temperature field is described by the input parameters vector exhaustively, the NN input is not the field itself but just the vector and timestep: $u_{\theta}(\lambda, t) = \hat{u}_{\lambda}(x,t)$. The NN architecture is a multilayer perceptron consisting of an input layer of 6 neurons, $L$ hidden layers of $H$ neurons with ReLU activation, and an output of $M^2$ neurons. It is trained using Adam optimizer with a learning rate of $1e^{-3}$. \hide{ and batch size $B$.}

In Melissa, we set the simultaneous job limit to  $m = 10$ and the \textit{Reservoir} watermark value to $300$, meaning that the NN training does not begin until the buffer contains at least $300$ unique samples. 

\subsection{Experiment orchestration}
\label{app:general-setup-tech-details}
To facilitate a broad analysis study, we employ \verb|Snakemake|\cite{snakemake}, a workflow management system that enables the execution and management of scalable, reproducible analysis studies. In our case, the workflow creates configuration files for Melissa runs across chosen grid.
To ensure the reproducibility\footnote{Code is available at \url{https://gitlab.inria.fr/melissa/ai4s-sc2024-heatpde.git}} of our experiments, we utilize \verb|Nix|~\cite{dolstra2004nix} package manager. 
The experiments were conducted on the Grid5000~\cite{grid} cluster using OAR scheduler~\cite{oar}.  Each  Melissa client as well as the  server run  one  48 processes MPI job, each one scheduled on a  48 core node.

\section{Experiments additional details}

\begin{figure}[H]
	\includegraphics[width=\columnwidth]{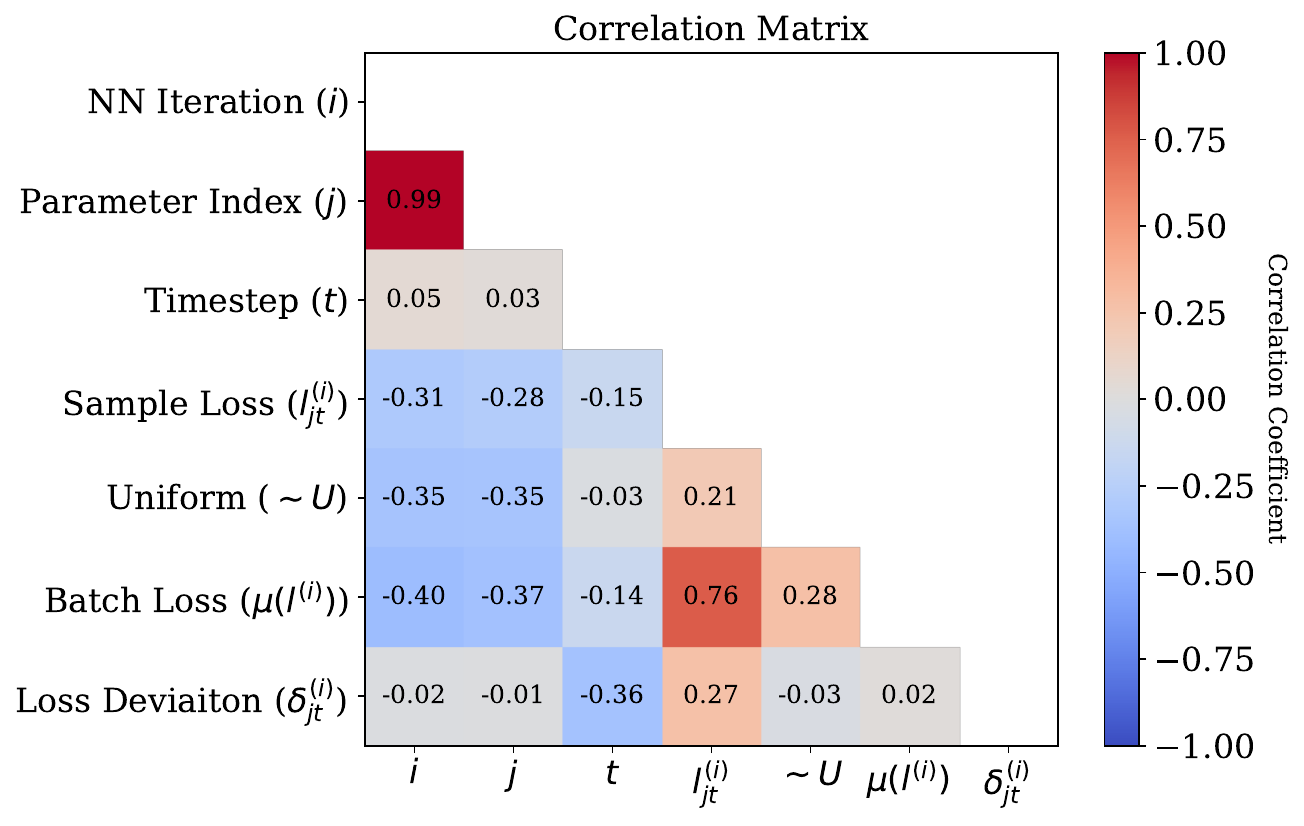}
	\caption{Correlation between per-sample and per-batch dynamics: indicated on the left axis. The upper triangle and diagonal values are omitted for readability. ``Uniform'' value is an indicator of whether the sample was produced by uniform mixing, ``loss deviation'' is the proposed metric, which is not dependent on NN iteration but still positively correlates with per-sample loss.}
	\label{fig:corr}
\end{figure}

Here we provide the Table~\ref{tab:fixed} with exact hyperparameters used in the study, and \ref{fig:corr} is the visualisation of correlation matrix.

\begin{table}[H]
	\caption{The fixed hyperparameters details according to varying (*) parameter: (1) the model size is varied, (2) $\sigma$ or $P$ or $N$ is varied, (3) $r_{s}$ or $r_{e}$ or $r_{c}$ is varied.}
	\label{tab:fixed}
	\begin{tabular}{l|l|l|l|l|l|l|l|l}
		& $\sigma$ & $P$   & $N$   & $r_s$ & $r_e$ & $r_c$ & $H$ & $L$ \\ \hline
		Study (1)           & 10.0     & 300   & 200   & 0.5   & 0.7   & 3     & *   & *   \\
		Study (2)     & */5.0    & */200 & */200 & 0.5   & 0.9   & 3     & 16  & 1   \\
		Study (3) & 5.0      & 200   & 200   & */0.1 & */1.0 & */5   & 16  & 1  
	\end{tabular}
\end{table}

\end{document}